\documentclass[11pt]{article}

\usepackage[final]{acl}

\usepackage{times}
\usepackage{latexsym}
\usepackage{amsmath}
\usepackage{amssymb}
\usepackage{graphicx}
\usepackage{booktabs} 
\usepackage{url}

\usepackage[table]{xcolor}
\usepackage{multirow}

\usepackage[T1]{fontenc}
\usepackage[utf8]{inputenc}
\usepackage{microtype}
\usepackage{inconsolata}

\usepackage{algorithm}        
\usepackage{algpseudocode}    

\title{Structural Anchor Pruning: Training-Free Multi-Vector Compression for Visual Document Retrieval}

\author{Zhuchenyang Liu, Ziyu Hu, Yao Zhang, Yu Xiao \\
  Aalto University \\
  Espoo, Finland \\
  \texttt{zhuchenyang.liu@aalto.fi}}

\begin{document}

\maketitle

\begin{abstract}
Recent Vision-Language Models (e.g., ColPali) enable fine-grained Visual Document Retrieval (VDR) but incur prohibitive multi-vector index storage overhead. Existing training-free pruning methods either rely on heuristic layer choices or degrade sharply under aggressive compression, leading prior work to argue that effective high-compression pruning requires query-dependent training. We challenge this view with Structural Anchor Pruning (SAP), a self-calibrating, training-free, query-agnostic index-time framework combining (i) Score Retention (SR), a white-box per-layer compression diagnostic; (ii) SR-guided window selection, which automatically locates the structural pruning region of any backbone with no per-model hyperparameters; and (iii) a visual in-degree centrality scorer that identifies anchor patches within that window. On ViDoRe v1/v2 across three architectures spanning 18, 28, and 36 backbone layers, SAP retains 93--96\% of NDCG@5 on v1 and 88--90\% on the harder v2 while pruning 90\% of visual tokens; at 20$\times$ compression it retains 85--90\% and 76--79\% respectively. Our layer-resolved SR analysis reveals an Alignment-Aggregation Divergence: visual structure is preserved as a stable ``Structural Plateau'' within the backbone, while the final layers reshape it into a sparse, query-aligned form unsuitable for pruning. Probing the pre-retrieval base backbones shows that contrastive fine-tuning sharpens this boundary three- to eight-fold, explaining why final-layer methods fail.\footnote{Code is available at \url{https://github.com/Ryenhails/structural-anchor-pruning}.}
\end{abstract}

\section{Introduction}
\label{sec:intro}

\begin{figure}[h]
    \centering
    \includegraphics[width=0.85\linewidth]{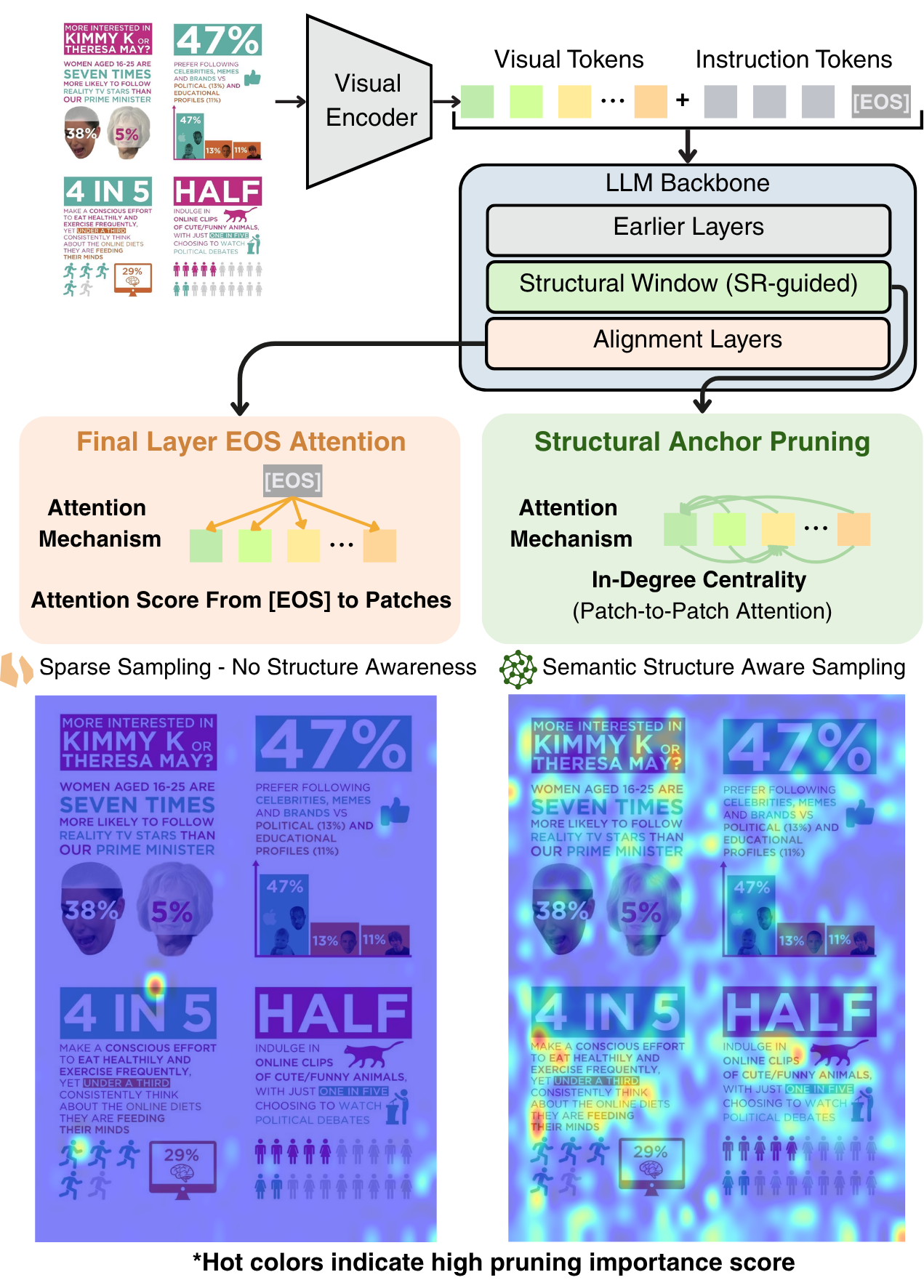}
    \caption{Comparison of Pruning Mechanisms.
    Final-layer EOS attention (left) yields sparse, structure-agnostic
    sampling. Our Structural Anchor Pruning (right) applies
    visual in-degree centrality on patch-to-patch attention
    within a structural window that is automatically located
    per architecture by our SR-guided selection procedure, yielding a structure-aware sample
    of anchor patches.}
    \label{fig:sap_comparison}
\end{figure}


\begin{figure*}[t]
    \centering
    \includegraphics[width=\textwidth]{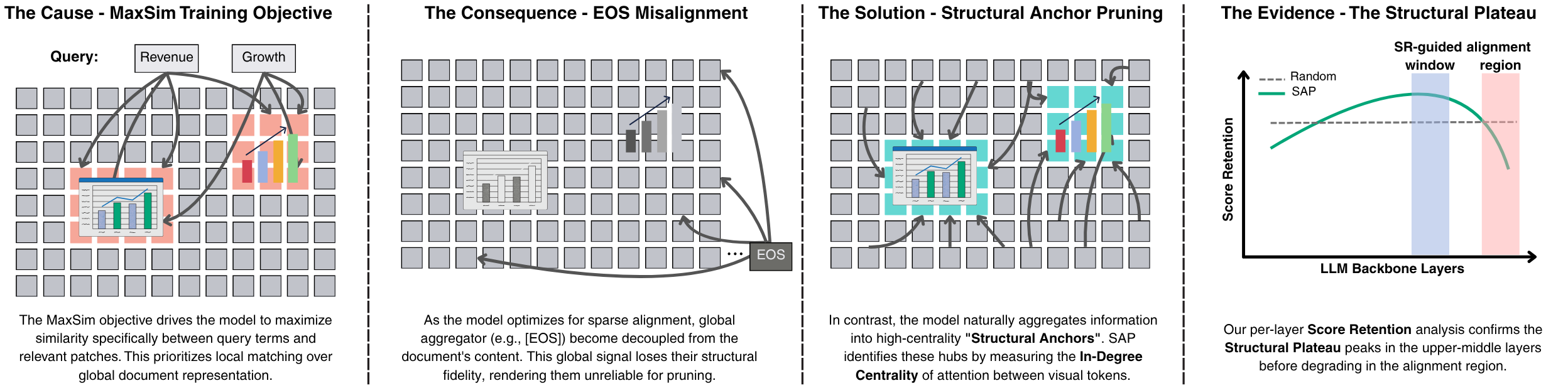}
    \caption{\textbf{The Alignment-Aggregation Divergence.}
    Final-layer global signals decay under MaxSim alignment, while a
    contiguous Structural Plateau within the backbone aggregates
    information into high-centrality anchor patches. SAP exploits this
    by measuring \textbf{visual in-degree centrality} between visual
    tokens; per-backbone Score Retention curves are reported in
    Figure~\ref{fig:subset_sweep}.}
    \label{fig:mechanism}
\end{figure*}

Visual Document Retrieval (VDR) has shifted from traditional pipelines to end-to-end Vision-Language Models (VLMs)~\cite{zhang2024visionlanguagemodelsvisiontasks}. VLM-based retrievers like ColPali~\cite{faysse2025colpaliefficientdocumentretrieval} achieve superior precision by representing documents as bags of visual patch embeddings. While this paradigm captures rich document structures through late interaction~\cite{khattab2020colbertefficienteffectivepassage}, it suffers from massive index size overhead. Addressing this scalability challenge through embedding compression is essential for deploying Visual RAG in realistic, large-scale scenarios.

To address this bottleneck, recent research has diverged into two primary streams. One involves training-based methods like Light-ColPali~\cite{ma-etal-2025-towards-storage} which are effective but require fully re-training and additional model modifications. Alternatively, training-free methods like DocPruner~\cite{yan2025docprunerstorageefficientframeworkmultivector} offer a modular solution, but these methods often suffer from performance degradation in high-compression regimes ($\ge 80\%$ reduction)~\cite{ma-etal-2025-towards-storage, yan2025docprunerstorageefficientframeworkmultivector}. Observing these challenges, ~\citet{ma-etal-2025-towards-storage} conclude that visual token importance for pruning is inherently query-dependent, thereby arguing that training-free pruning is insufficient for high compression ratio.


In this work, we challenge this conclusion. At the core of our approach is \textbf{Score Retention (SR)}, a white-box, label-free per-layer compression diagnostic that quantifies per-pair MaxSim fidelity decoupled from corpus-level ranking effects. We use SR as the driving signal of an \textbf{SR-guided window selection} procedure that automatically locates the structural pruning region for any backbone without per-model tuning. Within the selected window, \textbf{visual in-degree centrality} identifies anchor patches in a query-agnostic manner. We call the resulting end-to-end, training-free, query-agnostic, index-time framework \textbf{Structural Anchor Pruning (SAP)} (Figure~\ref{fig:sap_comparison}). On the ViDoRe v1/v2 benchmarks~\cite{faysse2025colpaliefficientdocumentretrieval, macé2025vidorebenchmarkv2raising}, SAP consistently outperforms EOS-Adaptive~\cite{yan2025docprunerstorageefficientframeworkmultivector}, Random~\cite{ma-etal-2025-towards-storage}, and Semantic Clustering~\cite{ma-etal-2025-towards-storage} across three backbones of widely different depth (18, 28, 36 layers), retaining 93--96\% of NDCG@5 on ViDoRe v1 and 88--90\% on ViDoRe v2 while reducing stored vectors by 90\%.

Beyond its operational role in window selection, SR also serves as a layer-resolved probe for analyzing VLM backbones. Applied across the full depth of three architectures, it reveals an \textbf{Alignment-Aggregation Divergence} (Figure~\ref{fig:mechanism}): a contiguous block of layers preserves the document's visual structure as a stable Structural Plateau, while the final layers reshape this representation into a sparse, retrieval-aligned form that is no longer suitable for pruning. This explains why SR-guided selection consistently lands at the boundary between the two regions, and why final-layer pruning methods (e.g., EOS-attention) degrade sharply under aggressive compression.

\section{Related Work and Preliminaries}
\label{sec:related_work}

\subsection{VDR Multi-Vector Late Interaction}
\label{sec:preliminaries}

Unlike dense retrievers that map a document to a single vector, multi-vector retrievers such as ColPali~\cite{faysse2025colpaliefficientdocumentretrieval} employ the \textbf{Multi-Vector Late Interaction} mechanism~\cite{khattab2020colbertefficienteffectivepassage}. A document image $D$ is encoded into a bag of visual patch embeddings $E_D = \{v_1, \dots, v_N\} \in \mathbb{R}^{N \times d}$ ($N \approx 1024$ patches per image). Given a text query $Q$ tokenized as $\{q_1, \dots, q_M\}$, the relevance score is computed via the MaxSim operator:
\begin{equation}
    S(Q, D) = \sum_{i=1}^{M} \max_{j=1}^{N} (q_i \cdot v_j)
\end{equation}
This preserves fine-grained layout details but requires indexing the full $E_D$, so storage scales linearly with $N$; on realistic corpora the index reaches terabytes~\cite{xu2025llamanemoretrievercolembedtopperforming}.

\subsection{Efficient Visual Document Retrieval}
\label{sec:efficient_vdr}

Two streams of work address this overhead: training-based adaptation and training-free pruning.

\noindent\textbf{Training-based Adaptation.} Methods like Light-ColPali~\cite{ma-etal-2025-towards-storage} employ knowledge distillation to train adapters that merge visual tokens into a smaller set of latent representations. While effective at high compression ratios, these approaches introduce significant operational overhead, requiring large-scale training data and full-model fine-tuning, which limits their zero-shot applicability to new architectures.

\noindent\textbf{Training-free Compression.} Conversely, training-free methods select a subset of informative patches $\hat{E}_D \subset E_D$ without model adaptation. Current strategies rely on one of four selection signals: (1) \textbf{Random Pruning}~\cite{ma-etal-2025-towards-storage} assumes a holographic distribution of visual information and selects patches via uniform pruning; (2) \textbf{Semantic Clustering}~\cite{ma-etal-2025-towards-storage} aims to reduce redundancy by grouping embeddings via K-Means and indexing only the cluster centroids; (3) \textbf{EOS-Attention}~\cite{ma-etal-2025-towards-storage} selects patches based on their cross-attention weights with the final \texttt{[EOS]} token; and (4) \textbf{EOS-Adaptive Pruning (DocPruner)}~\cite{yan2025docprunerstorageefficientframeworkmultivector} extends this by dynamically adjusting the pruning ratio based on information density. However, these training-free methods suffer from severe performance degradation when pushed to high compression ratios (e.g., $>80\%$ reduction)~\cite{yan2025docprunerstorageefficientframeworkmultivector, ma-etal-2025-towards-storage}. Consequently, they argue that static, query-agnostic pruning strategies are insufficient for high ratio compression~\cite{ma-etal-2025-towards-storage}. Closely related training-free efforts include hierarchical patch compression with dynamic pruning and quantization~\cite{bach2025hierarchicalpatchcompressioncolpali} and training-free pooling within a multi-stage search pipeline~\cite{yeroyan2026visualragtoolkitscaling}; both compress the index without adaptation, but neither localizes the layer at which the structural signal is read.

Concurrent training-free work introduces a Prune-then-Merge framework~\cite{yan-etal-2026-sculpting}, addressing the information loss of pure pruning with a two-stage pipeline: adaptive pruning is followed by hierarchical merging that aggregates the discarded patches into summary representations. SAP is procedurally simpler, a single-stage, fixed-budget per-document top-$k$ selection driven by attention-graph centrality read from a backbone window automatically located by SR. The two designs target different stages of the indexing pipeline and could in principle be stacked.

\noindent\textbf{Inference-time vs.\ Index-time Pruning.}
A separate line of work targets visual token pruning at inference time to accelerate single-pass VLM forward latency, such as Token Merging~\cite{bolya2023tokenmergingvitfaster} for vision transformers and SparseVLM~\cite{zhang2025sparsevlmvisualtokensparsification} for vision-language models. These methods drop or merge tokens dynamically during the forward pass and discard them after inference. SAP addresses a different setting: persistent index-time compression that prunes once and reuses the result query-agnostically across all future queries, optimizing storage rather than per-pass latency. We therefore compare only against index-time baselines.

\begin{figure*}[t]
    \centering
    \includegraphics[width=\textwidth]{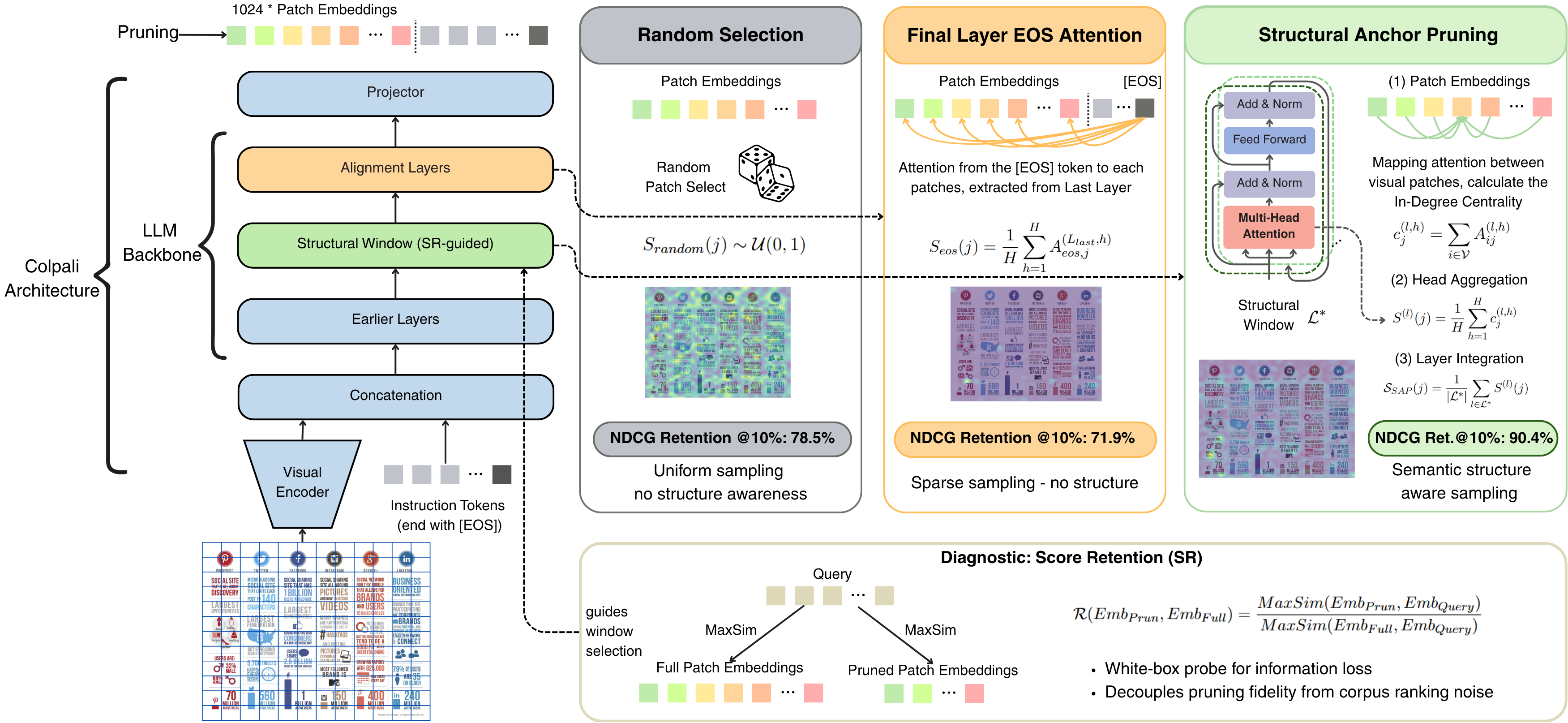}
    \caption{\textbf{Overview of SAP.} We compare three pruning paradigms on the ColPali architecture. \textbf{Left:} the shared Vision-Language backbone processes the image. \textbf{Middle:} conventional methods (Random Selection, Final Layer EOS Attention) fail to identify critical tokens, resulting in lower retention. \textbf{Right:} SAP identifies semantic structural anchor patches via \textbf{visual in-degree centrality} within the \textbf{Structural Window} of the backbone, automatically located by our SR-guided procedure.
    \textbf{Bottom:} the \textbf{Score Retention (SR)} diagnostic directly compares the MaxSim scores of pruned versus full embeddings, isolating intrinsic information loss from corpus-dependent ranking noise; SR itself drives the Structural Window selection.
    NDCG@5 retention numbers shown are ColPali averages on ViDoRe v2 at $\gamma=0.10$ (Table~\ref{tab:global_results}).}
    \label{fig:method_overview}
\end{figure*}

\section{Methodology}
\label{sec:methodology}

We first introduce Structural Anchor Pruning, which scores tokens by visual in-degree centrality within a structural window of the LLM backbone. We then introduce Score Retention (SR), a white-box compression diagnostic that quantifies per-pair MaxSim fidelity independently of corpus-level ranking. Finally, we present SR-guided window selection, a label-free procedure that uses SR to automatically locate the structural window for any backbone, eliminating per-model tuning. The overall framework is illustrated in Figure~\ref{fig:method_overview}.

\subsection{Structural Anchor Pruning}
\label{SAP_method}

We propose SAP, a training-free strategy designed to extract the intrinsic semantic structure of a document image. We identify semantic structural anchor patches by measuring the visual In-Degree Centrality of tokens within the Large Language Model (LLM) backbone. We hypothesize that semantic structural patches in the backbone's Structural Plateau act as information hubs, aggregating features from numerous other regions and constituting the core semantic representation of the document.

We treat the self-attention mechanism within the LLM layers at any given layer $l$ as a directed graph, where nodes represent image patches and edges represent attention weights. Mechanistically, the attention weight $A_{ij}$ represents the importance score of token $j$ for token $i$. Consequently, the summation over all source indices, $\sum_{i} A_{ij}$, quantifies the total importance of token $j$ across all visual tokens, serving as a direct proxy for its global influence.

To isolate the visual structure, we restrict our calculation to the \textbf{Visual-to-Visual} attention, masking out attention scores involving text tokens (e.g., system prompts). Let $A^{(l, h)} \in \mathbb{R}^{T \times T}$ be the full attention matrix for head $h$ over sequence length $T$, and $\mathcal{V}$ be the set of indices corresponding to visual patches. The importance of a visual patch $j \in \mathcal{V}$ at layer $l$ is defined by its column-sum:
\begin{equation}
    c^{(l, h)}_j = \sum_{i \in \mathcal{V}} A^{(l, h)}_{ij}
\end{equation}
A high in-degree indicates that patch $j$ acts as a central aggregator within the visual modality.

\paragraph{Head Aggregation.}
To synthesize signals across the $H$ attention heads at layer $l$, we average centralities across heads, which prioritizes anchors that are consistently active across attention subspaces:
\begin{equation}
    S^{(l)}(j) = \frac{1}{H} \sum_{h=1}^{H} c^{(l, h)}_j
\end{equation}

\paragraph{Layer Integration.}
Standard pruning approaches typically rely on the final layer ($l=L_{total}$) under the assumption that it represents the most refined semantic state. However, we hypothesize a two-phase structure within the backbone: an aggregation phase, where tokens actively exchange information to build a cohesive structural understanding of the document, followed by an alignment phase in the final layers, where representations are implicitly reorganized to optimize the contrastive retrieval objective (MaxSim). This final alignment often ``sparsifies'' the attention map to fit potential query distributions, thereby degrading the intrinsic structural signals required for effective pruning.

To capture the robust structural core before this degradation occurs, we introduce Layer Integration. We define the layer ensemble $\mathcal{L}^*$ as a function of the model's total depth $L_{total}$ and relative depth boundaries $\alpha, \beta \in [0, 1]$:
\begin{equation}
\label{eq:layer_ensemble}
    \mathcal{L}^*(\alpha, \beta) = \{ l \in \mathbb{N} \mid \lfloor \alpha \cdot L_{total} \rfloor \le l < \lfloor \beta \cdot L_{total} \rfloor \}
\end{equation}
where $0 \le \alpha < \beta \le 1$ define the boundaries of the structural window. We delegate the choice of $(\alpha, \beta)$ to a label-free, automatic procedure introduced in Section~\ref{sec:osr_window}. The final importance score $\mathcal{S}_{SAP}(j)$ is obtained by averaging centrality scores across this window:
\begin{equation}
    \mathcal{S}_{SAP}(j) = \frac{1}{|\mathcal{L}^*|} \sum_{l \in \mathcal{L}^*} S^{(l)}(j)
\end{equation}

\subsection{Score Retention}
\label{sec:oracle_protocol}

While standard evaluation metrics like Normalized Discounted Cumulative Gain (NDCG)~\cite{wang2013theoreticalanalysisndcgtype} are essential for assessing retrieval effectiveness, they are insufficient for diagnosing the intrinsic fidelity of pruned representations. Formally, NDCG at position $k$ is defined as:
\begin{equation}
    \mathrm{NDCG}@k = \frac{1}{\mathrm{IDCG}_k} \sum_{i=1}^{k} \frac{2^{rel_i} - 1}{\log_2(i+1)}
\end{equation}
where $rel_i$ is the relevance score, $i$ is the ranking position, and $\mathrm{IDCG}_k$ is the Ideal Discounted Cumulative Gain, acting as a normalization factor to ensure the score lies in $[0,1]$.
Crucially, the logarithmic term $\log_2(i+1)$ explicitly couples the evaluation to the relative rank $i$. This makes the metric inherently corpus-dependent: a drop in NDCG may result from the presence of hard negatives shifting the rank $i$, rather than a loss of information in the document representation itself. Consequently, when used to measure pruning performance, NDCG confounds pure information loss with the model's discriminative capacity.

To disentangle these factors and isolate the intrinsic visual information retained by a pruning method, we introduce \textbf{Score Retention (SR)} as a white-box compression diagnostic. We define fidelity not by ranking position, but by the preservation of the raw MaxSim score:
\begin{equation}
\label{eq:sr}
    \mathcal{R}(\hat{E}_D, E_D) = \frac{\sum_{i=1}^M \max_{v \in \hat{E}_D} (q_i \cdot v)}{\sum_{i=1}^M \max_{v \in E_D} (q_i \cdot v)}
\end{equation}
A retention of $1.0$ confirms that the pruned patches $\hat{E}_D$ preserve the exact visual features triggered by the query, independent of the document's ranking relative to distractors.

SR is intentionally distinct from NDCG. Whereas NDCG measures corpus-level ranking quality relative to gold relevance judgments, SR measures per-pair score fidelity for individual (query, document) pairs, decoupled from any corpus-level ranking effect. The two are complementary, not interchangeable: a high SR does not guarantee correct ranking against distractors, and a low SR can still leave rankings unchanged. SR is purpose-built for one specific question: which layers preserve the most score fidelity under a fixed compression budget? We use it in two ways: (i) as the driving signal for our automatic window-selection procedure, and (ii) as a layer-resolved probe to localize where structural information concentrates within the backbone.

\subsection{SR-Guided Window Selection}
\label{sec:osr_window}

A naive instantiation of SAP would require selecting the structural window $(\alpha, \beta)$ heuristically. We propose a label-free, training-free procedure that uses SR itself to locate the window for any backbone, parameterized only by a calibration set size $N$ and a relative window width $\rho \in (0, 1]$.

The procedure operates as follows. Given an unlabeled calibration set $\mathcal{C} = \{(I_n, Q_n)\}_{n=1}^N$ and a target retention ratio $\gamma$, for each layer $l \in \{0, \dots, L_{total}-1\}$ we compute, for every pair, the centrality $S^{(l)}$ from layer $l$ alone (Eq.~3), retain the top-$\lceil \gamma |E_I| \rceil$ patches of the image embeddings, and record SR (Eq.~\ref{eq:sr}) between the pruned and full embeddings. Averaging across calibration pairs yields a per-layer score retention curve $\mathcal{R}(l)$. We compute the median $m = \mathrm{median}\{\mathcal{R}(l)\}_{l=0}^{L_{total}-1}$ and identify the alignment region as the longest suffix $[l^*,\, L_{total}-1]$ for which $\mathcal{R}(l) < m$ throughout, i.e., the contiguous tail where retention has fallen below the global median. The structural window is placed immediately before this region, with width $k = \lceil \rho \cdot L_{total} \rceil$ layers:
\begin{equation}
\beta = \frac{l^*}{L_{total}}, \qquad \alpha = \frac{l^* - k}{L_{total}}.
\end{equation}
The window is never empty: $k = \lceil \rho \, L_{total} \rceil \ge 1$ for any
$\rho \in (0, 1]$, and $\alpha$ is clamped at $0$ whenever $l^* < k$, so
$\mathcal{L}^*(\alpha, \beta)$ always contains at least one layer.

Formal pseudocode is provided in Appendix~\ref{app:osr_algorithm}.

The procedure has three properties. It requires no labels, no training, and no NDCG evaluation. It is identical across backbones, introducing no per-architecture hyperparameter. And it directly operationalizes the Alignment-Aggregation Divergence (Section~\ref{sec:empirical_validation}) by placing the window at the empirically detected boundary between aggregation and alignment phases.

The calibration queries are unlabeled and disjoint from the deployment query stream: they may come from any held-out generic query source and are used only once, offline. Once $(\alpha, \beta)$ has been selected, the window is applied query-agnostically at index time: SAP's centrality scorer never sees a query, neither at calibration nor at deployment.

\begin{figure*}[t]
    \centering
    \includegraphics[width=\textwidth]{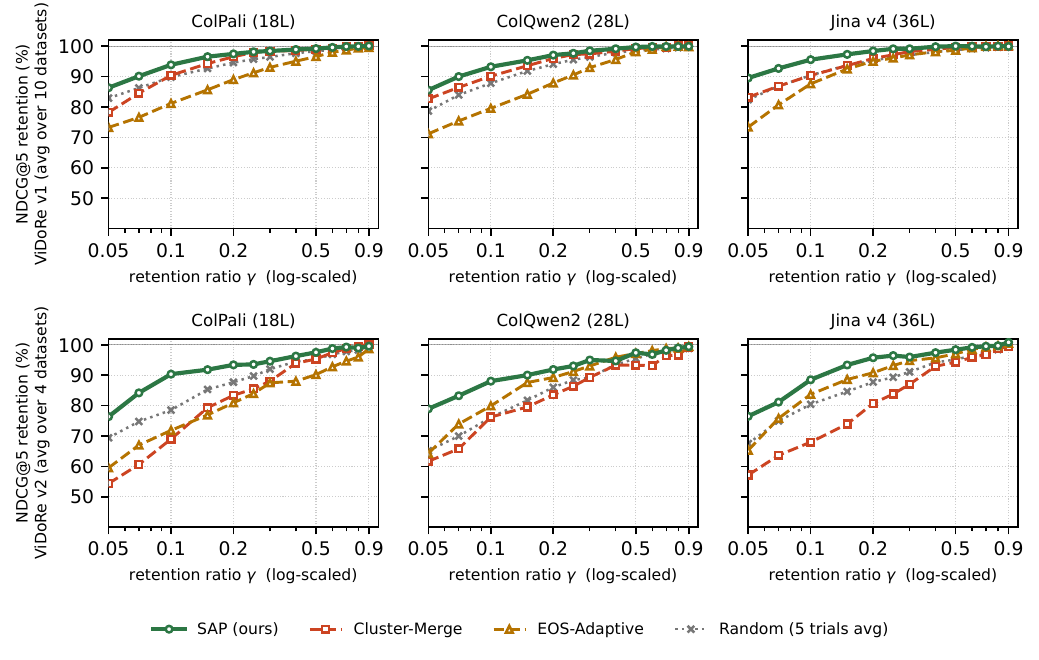}
    \caption{\textbf{Efficiency-Fidelity Trade-off.} NDCG@5 retention vs.\ retention ratio $\gamma$ across ColPali, ColQwen2, and Jina v4 on ViDoRe v1 (top) and v2 (bottom). SAP (green) consistently dominates training-free baselines at every $\gamma$ on every backbone-benchmark pair, with the largest gains under aggressive compression ($\gamma \le 0.10$). SAP uses the SR-guided window.}
    \label{fig:retention_curve}
\end{figure*}

\section{Evaluation}
\label{sec:comprehensive_eval}

In this section, we comprehensively benchmark the performance of SAP on large-scale visual retrieval tasks. We evaluate SAP's ability to maintain high retrieval fidelity across diverse architectures and datasets, compare it against state-of-the-art training-free and training-based baselines, and assess its computational efficiency.

\subsection{Evaluation Setup}

We employ three distinct VLM architectures to evaluate SAP: \textbf{ColPali} (SigLIP + PaliGemma)~\cite{beyer2024paligemmaversatile3bvlm, faysse2025colpaliefficientdocumentretrieval}, representing the standard fixed-patch retrieval paradigm, and \textbf{ColQwen2} (NaViT + Qwen2-VL)~\cite{wang2024qwen2vlenhancingvisionlanguagemodels, faysse2025colpaliefficientdocumentretrieval}, representing a dynamic-resolution architecture with a deeper backbone. We also extend our evaluation to a SOTA model \textbf{Jina Embeddings v4} with Qwen2.5-VL backbone~\cite{günther2025jinaembeddingsv4universalembeddingsmultimodal, bai2025qwen25vltechnicalreport}. Incorporating these architectures allows us to assess the generalizability of SAP across different VLM backbones and distinct embedding optimization recipes. See Appendix \ref{app:model_details} for model architectural specifications.

We instantiate the layer ensemble $\mathcal{L}^*$ (Eq.~\ref{eq:layer_ensemble}) automatically using the SR-guided window selection procedure, with hyperparameters set to $\rho = 0.2$ (relative window width) and $N = 500$ (calibration set size). The calibration set consists of 500 (image, query) pairs uniformly sampled from the ColPali training corpus~\cite{faysse2025colpaliefficientdocumentretrieval}, which is disjoint from all ViDoRe v1 and v2 evaluation splits. The resulting per-architecture windows are listed in Table~\ref{tab:layer_instantiation}. This calibration runs once per backbone, offline, and the pruned indices are reused for all future user queries with no further re-calibration.

We compare our SAP against three distinct training-free pruning paradigms (details in Appendix~\ref{app:baseline_details}): (1) Adaptive-EOS: An EOS attention-based method proposed by DocPruner \cite{yan2025docprunerstorageefficientframeworkmultivector}, which employs document-specific thresholding based on final-layer global (\texttt{[EOS]}) attention scores. To ensure fair comparison, we apply a quantile-based calibration to align its global retention rate strictly with our fixed-ratio methods; (2) Random: The robust stochastic pruning baseline~\cite{ma-etal-2025-towards-storage}; and (3) Semantic Cluster: K-Means clustering on final embeddings, identified by ~\citet{ma-etal-2025-towards-storage} as the state-of-the-art training-free compression approach.

We utilize the full \textbf{ViDoRe v1}~\cite{faysse2025colpaliefficientdocumentretrieval} and \textbf{ViDoRe v2}~\cite{macé2025vidorebenchmarkv2raising} benchmarks. These cover a wide spectrum of domains. Detailed dataset statistics are provided in Appendix \ref{app:dataset_details}.

\subsection{Main Results}
\label{sec:main_results}

Figure~\ref{fig:retention_curve} reports NDCG@5 retention as a function of the retention ratio $\gamma$ for all three backbones on both ViDoRe v1 and v2. Three observations emerge across all six panels.

The SAP curve (green) matches or exceeds every training-free baseline at every $\gamma \in [0.05, 0.9]$ on every (backbone, benchmark) pair. At high retention ratios ($\gamma \ge 0.5$) all methods cluster near $95$--$100\%$ of full-model NDCG@5 and the differences are small. The separation widens monotonically as compression intensifies, and by $\gamma=0.05$ baselines fan out across a $\sim$30-point band while SAP remains comfortably on top.

At the practical deployment regime, SAP retains 93--96\% of NDCG@5 on ViDoRe v1 and 88--90\% on the harder v2 at $\gamma=0.10$, and 85--90\% and 76--79\% respectively at $\gamma=0.05$, while Adaptive-EOS and Cluster-Merge fall to 54--65\% on v2. On v2 at $\gamma=0.05$, SAP leads every training-free baseline on every backbone, by $7$--$14$ NDCG@5 points over the strongest baseline and by up to $22$ points over the weakest, closing the gap that prior work~\cite{ma-etal-2025-towards-storage,yan2025docprunerstorageefficientframeworkmultivector} cited as the principal limitation of training-free pruning.

The SAP curve shape is qualitatively identical across backbones of widely different depth (18, 28, 36 layers), confirming that the SR-guided procedure  tracks the optimal window per architecture without any per-model tuning. Appendix~\ref{app:three_gamma} provides a three-$\gamma$ case study with raw NDCG@5 values and Score Retention, and Appendix~\ref{app:ndcg_cutoffs} reports per-cutoff (NDCG@$\{1,5,10,50,100\}$) retention.

\subsection{SR-Guided Window Validation}
\label{sec:window_ablation}

We validate the SR-guided window selection procedure (Section~\ref{sec:osr_window}) by comparing it against a brute-force search over 9 sliding windows of fixed width $0.2\,L_{total}$, stepped by $0.1\,L_{total}$, covering relative depth ranges from 0--20\% to 80--100\%. Table~\ref{tab:window_ablation} reports NDCG@5 retention at $\gamma=0.10$ on ViDoRe v1 and v2 for each of the 9 windows on all three backbones. Full results for $\gamma=0.20$ and $\gamma=0.05$ are deferred to Appendix~\ref{app:full_window_ablation}.

The SR-guided procedure selects the 60--80\% window for ColPali and ColQwen2, and the 70--90\% window for Jina v4 (bold cells in the Mean rows of Table~\ref{tab:window_ablation}). On the mean of ViDoRe v1 and v2, this selection coincides exactly with the brute-force-best window for all three backbones. Per-bench, ColQwen2 matches the brute-force best on both v1 and v2; Jina v4 matches the best on v1 and lies $0.2$ NDCG@5 points behind the 50--70\% best on v2; ColPali matches the best on v2 and lies $0.3$ points behind the 50--70\% best on v1. The largest per-bench gap to the brute-force optimum across the three backbones is therefore $0.3$ NDCG@5, achieved without any retrieval labels (Table~\ref{tab:layer_instantiation}, Appendix~\ref{app:layer_instantiation}).

The brute-force optimum lies at a different relative depth on each backbone: 50--70\% for ColPali (18 layers), 60--80\% for ColQwen2 (28 layers), and 70--90\% for Jina v4 (36 layers). In absolute terms, the optimum always ends approximately 2--4 layers before the last layer, regardless of backbone depth. The SR-guided procedure recovers this shifted optimum on all three backbones without any per-model intervention; we examine the mechanism behind the shift in Section~\ref{sec:empirical_validation}.

Holding the window position fixed by the drop-alignment rule and sweeping the width $k$ from $1$ to $L_{total}$ layers shows NDCG@5 retention is insensitive to $k$ over a broad plateau that includes our default $k = \lceil 0.2\,L_{total} \rceil$ for all three backbones (Figure~\ref{fig:k_robustness}, Appendix~\ref{app:k_robustness}). SAP is therefore robust to moderate deviations from the default window width.

\begin{table}[t]
\centering
\small
\setlength{\tabcolsep}{2.8pt}
\caption{\textbf{Brute-force window ablation at $\gamma=0.10$.}
NDCG@5 retention (\%) for SAP at nine sliding windows of fixed width
$0.2\,L_{total}$, stepped by $0.1\,L_{total}$ across the LLM backbone.
$^\dagger$ marks the window selected by our SR-guided procedure; the brute-force best window per row is in
\textbf{bold}. The SR-guided window coincides with the bold-and-dagger
cell in each Mean row, i.e., it is optimal on the v1+v2 mean for all
three backbones. Full results at $\gamma=0.20$ and $\gamma=0.05$ in
Appendix~\ref{app:full_window_ablation}.}
\label{tab:window_ablation}
\resizebox{\columnwidth}{!}{
\begin{tabular}{l l c c c c c c c c c}
\toprule
\textbf{Model} & \textbf{Bench}
& 0--20 & 10--30 & 20--40 & 30--50 & 40--60 & 50--70 & 60--80 & 70--90 & 80--100 \\
\midrule
\multirow{3}{*}{ColPali}
& v1 & 88.2 & 91.8 & 92.6 & 93.1 & 92.7 & \textbf{94.2} & 93.9$^\dagger$ & 91.9 & 90.0 \\
& v2 & 82.9 & 86.8 & 87.9 & 90.6 & 89.9 & 88.3 & \textbf{90.9}$^\dagger$ & 85.1 & 81.5 \\
& \textit{Mean} & 85.6 & 89.3 & 90.2 & 91.9 & 91.3 & 91.3 & \textbf{92.4}$^\dagger$ & 88.5 & 85.7 \\
\cmidrule(lr){1-11}
\multirow{3}{*}{ColQwen2}
& v1 & 85.2 & 88.6 & 90.0 & 90.7 & 91.3 & 92.0 & \textbf{93.3}$^\dagger$ & 92.2 & 91.5 \\
& v2 & 82.6 & 82.7 & 82.1 & 82.0 & 81.4 & 84.3 & \textbf{88.5}$^\dagger$ & 85.2 & 80.9 \\
& \textit{Mean} & 83.9 & 85.6 & 86.1 & 86.3 & 86.4 & 88.1 & \textbf{90.9}$^\dagger$ & 88.7 & 86.2 \\
\cmidrule(lr){1-11}
\multirow{3}{*}{Jina v4}
& v1 & 88.3 & 89.8 & 91.3 & 91.5 & 92.1 & 93.6 & 94.9 & \textbf{95.6}$^\dagger$ & 94.6 \\
& v2 & 82.4 & 81.7 & 83.6 & 87.8 & 88.5 & \textbf{88.7} & 87.5 & 88.5$^\dagger$ & 85.6 \\
& \textit{Mean} & 85.3 & 85.7 & 87.5 & 89.7 & 90.3 & 91.1 & 91.2 & \textbf{92.0}$^\dagger$ & 90.1 \\
\bottomrule
\end{tabular}
}
\end{table}

\begin{figure*}[ht]
    \centering
    \includegraphics[width=\textwidth]{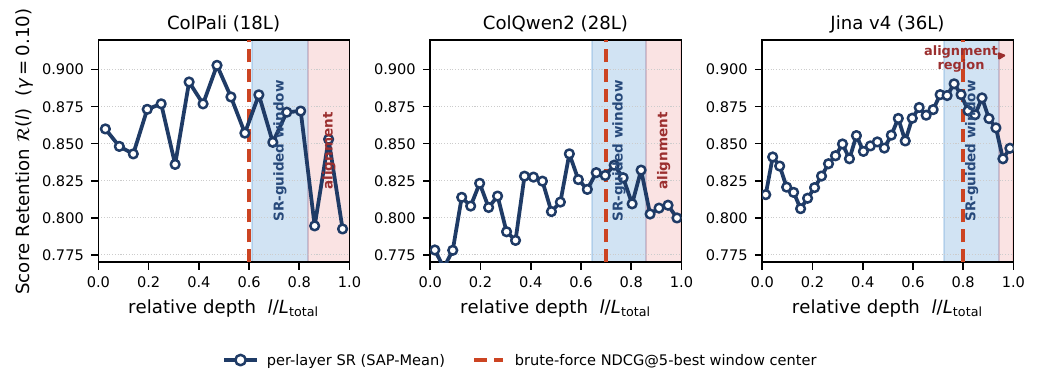}
    \caption{\textbf{Per-layer Score Retention and SR-guided window selection across three backbones.} For each backbone, $\mathcal{R}(l)$ at $\gamma=0.10$ is averaged over the 500-pair calibration set. The blue band marks the window selected by our SR-guided procedure; the red band marks the alignment region; the vertical dashed line marks the brute-force NDCG@5-best window center (Table~\ref{tab:window_ablation}). The two window agree to within $\sim$1\% NDCG@5. The Alignment-Aggregation Divergence is visible in all three backbones as a structural plateau followed by a sharp suffix drop in the alignment region.}
    \label{fig:subset_sweep}
\end{figure*}

\subsection{Computational Efficiency}

Beyond retrieval fidelity, SAP maintains high operational throughput. Theoretical complexity analysis and empirical benchmarks (Appendix~\ref{sec:empirical_latency}) show that SAP adds negligible overhead to the total forward pass latency. In contrast, the clustering-based method incurs a significant $6\%$ computational overhead.

On the ViDoRe~v2 union ($3{,}006$ documents, $1{,}152$ queries), SAP at $\gamma=0.10$ shrinks the ColPali index from $751$~MB to $75$~MB ($10\times$ reduction) and accelerates end-to-end MaxSim retrieval from $6.1$~ms to $0.78$~ms per query ($7.9\times$ speedup), while retaining 90.4\% of NDCG@5. At $\gamma=0.05$ the index drops to $37$~MB ($20\times$) and latency to $0.47$~ms ($13\times$). The full benchmark across all $\gamma$ values is provided in Appendix~\ref{app:storage_latency}.

\subsection{Comparison with Trained Method}
\label{sec:comparison_trained}

We also benchmark SAP against \textbf{Light-ColPali}~\cite{ma-etal-2025-towards-storage}, a trained token-merging baseline. Light-ColPali's numbers are quoted from its own paper, where they are averaged over a different evaluation set and normalized by a different full-model upper bound, so this is an indicative comparison (see Appendix~\ref{app:trained_comparison}). Read as retention against each method's own upper bound, at moderate compression ($4\times$, $\gamma=0.25$) training-free SAP is within $0.2$--$1.3$ points of the trained method; at $9\times$ ($\gamma=0.10$) it trails by $3.9$--$5.0$ points; and at extreme $25\times$ ($\gamma=0.05$) the gap widens to $8.9$--$10.7$ points in Light-ColPali's favor, consistent with the expected advantage of learned feature fusion in the most aggressive regime. The full comparison table is provided in Appendix~\ref{app:trained_comparison}.

\subsection{Score Retention vs.\ NDCG}
\label{sec:sr_ndcg_analysis}

We assess the empirical relationship between SR and downstream NDCG. Across all (model, benchmark, method, $\gamma$, dataset) configurations we observe a moderate positive Pearson correlation ($r \approx 0.60$). This aligns with what theory predicts: SR measures per-pair score fidelity, NDCG measures corpus-level ranking quality relative to hard negatives, and these are conceptually distinct quantities. The full SR--NDCG scatter is in Appendix~\ref{app:sr_ndcg_scatter}.

\section{Alignment-Aggregation Divergence}
\label{sec:empirical_validation}

Having established the retrieval performance of SAP, we turn to the mechanistic question: why does the semantic signal needed for pruning decouple from the final retrieval embedding? We apply the SR protocol to every layer of the three backbones using the same calibration set from evaluation. Figure~\ref{fig:subset_sweep} reveals two distinct phases in every backbone, regardless of depth.

Before the alignment region, the per-layer SR forms a sustained plateau (blue band in Figure~\ref{fig:subset_sweep}). Attention here aggregates local visual features into high-centrality anchor patches that constitute the document's ``semantic core''. The plateau appears consistently across all three evaluated backbones despite a 2$\times$ depth range from 18 to 36 layers.

Approaching the final layers, SR drops sharply (red band). We attribute this to the contrastive MaxSim objective: to maximize separability against hard negatives, the model reorganizes representations into a sparse, query-aligned form. This optimization is beneficial for ranking but destroys the dense structural signal needed for pruning. Adaptive-EOS accordingly falls at or below Random on ViDoRe~v1 across all three backbones, and on both benchmarks under extreme compression ($\gamma=0.05$); at moderate ratios on ViDoRe~v2 the two are mixed (Table~\ref{tab:global_results}). In every regime it stays well below SAP. The empirically detected boundary between the two phases is precisely what the SR-guided window selection procedure operationalizes.

\subsection{Is the Divergence Induced by Retrieval Fine-Tuning?}
\label{sec:base_probe}

The account above attributes the late-layer SR collapse to the contrastive
retrieval objective. That attribution is testable directly: if the collapse
were an intrinsic property of deep VLMs, the base backbones, before any
retrieval fine-tuning, should exhibit it too.

We repeat the per-layer SR protocol with a single variable
changed, the source of the attention. For ColPali we read attention from PaliGemma-3B, and for ColQwen2 from Qwen2-VL-2B-Instruct; these are the pre-retrieval checkpoints the two retrievers are built on. Centrality (Eq.~2--3) is computed from those
attentions, the retrieval model's embeddings are pruned to the top
$\gamma N$ patches under that ranking, and SR (Eq.~\ref{eq:sr}) is recorded per
layer. Images, queries, embedding space, calibration set (the same 500 pairs,
seed 42) and budget are identical, and token positions are aligned because both
retrievers reuse their base model's processor and patch layout. We omit Jina~v4,
whose released checkpoint has no separately available pre-retrieval model;
the claim below is therefore bounded to ColPali and ColQwen2.

\begin{figure}[t]
    \centering
    \includegraphics[width=\columnwidth]{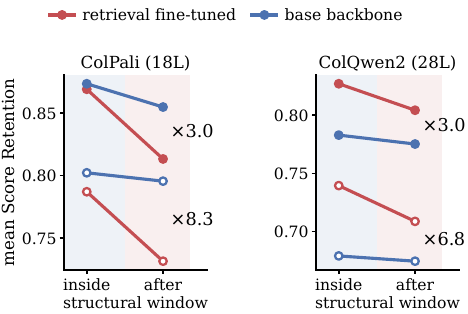}
    \caption{\textbf{Retrieval fine-tuning sharpens the plateau/alignment
    boundary rather than creating it.} Mean Score Retention inside the SR-guided
    structural window versus after it, with
    attention read from the retrieval-fine-tuned model (red) or from its
    base backbone (blue); filled markers $\gamma=0.10$, hollow
    markers $\gamma=0.05$.}
    \label{fig:base_probe}
\end{figure}

Figure~\ref{fig:base_probe} shows the two attention sources side by side.
We summarize each curve by the mean SR inside the SR-guided window and the mean
SR over the alignment region that follows it. Both quantities are tabulated for all four
settings in Appendix~\ref{app:base_probe_values}. 

Pruning driven by base-model attention shows a mild late-layer decline of its own, so fine-tuning does not create the effect from nothing. The step across the window boundary is nonetheless 3$\times$ to 8$\times$ larger when the attention comes from the fine-tuned model, and the factor grows as the compression budget tightens from $\gamma=0.10$ to $\gamma=0.05$.
We therefore state the mechanism in its evidence-bounded form: on these two backbones, contrastive retrieval fine-tuning does not introduce the plateau/alignment split, but sharpens it into the reliably locatable boundary that the SR-guided procedure keys on.

\section{Conclusion}
\label{sec:conclusion}

We propose Structural Anchor Pruning (SAP), a self-calibrating, training-free, and query-agnostic index-time compression framework for multi-vector visual document retrieval. SAP couples a white-box Score Retention (SR) diagnostic with an SR-guided window selection procedure that automatically locates the structural pruning region of any backbone without labels or per-model tuning, and uses visual in-degree centrality to identify anchor patches within that window. SAP reduces index storage by $90\%$ while retaining 93--96\% of NDCG@5 on ViDoRe v1 and 88--90\% on ViDoRe v2, across three architectures spanning 18 to 36 layers. Our layer-resolved analysis uncovers the Alignment-Aggregation Divergence: a stable Structural Plateau within the backbone preserves the document's visual structure, while the final layers compress this representation into a sparse, retrieval-aligned form that is no longer suitable for pruning. A probe against the pre-retrieval base backbones indicates that, on ColPali and ColQwen2, contrastive retrieval fine-tuning does not create this divergence but sharpens it three- to eight-fold, explaining why final-layer
methods fail.

\section*{Limitations}
\label{sec:limitations}

Our evaluation is confined to the multi-vector late-interaction VDR paradigm. Applicability to single-vector dense retrievers, generic image-text matching, and end-to-end retrieval-augmented generation pipelines is unexplored.

The calibration set is drawn from a VDR-relevant corpus (the ColPali training data), and our per-layer SR analysis is averaged over this distribution. The Structural Plateau location and the SR-guided window may therefore shift under more aggressive visual-domain changes, such as natural-image retrieval, scientific figures, or satellite imagery, where the visual statistics and the relevant feature scales differ markedly from document pages. Verifying domain robustness under such shifts is left for future work.

The evidence we offer for calibration robustness is bounded to the document-image domain: it covers two disjoint calibration distributions (Appendix~\ref{app:calib_robustness}) and calibration sizes from 50 to 1000 pairs, but no domain outside visual document retrieval. We claim stability within that scope only, not domain universality.

Our account of the mechanism is likewise bounded. The base-backbone probe (Section~\ref{sec:base_probe}) covers ColPali and ColQwen2, the two backbones whose pre-retrieval checkpoints are separately available; whether the same sharpening holds for Jina~v4 is untested. The probe is also observational rather than interventional: it contrasts two checkpoints, and does not isolate which part of the retrieval fine-tuning recipe is responsible.

\section*{Acknowledgments}

This work was carried out in AiWo: Human-Centric, AI-Enabled Collaborative Fieldwork
Operations, a Business Finland co-innovation programme under grant agreement No.~69/31/2025.
The computations were performed using computer resources provided by the Aalto University
School of Science ``Science-IT'' project.

We used Claude Code as a coding and writing assistant: implementing coding scripts, preparing and analyzing data, and assisting with the language of
the manuscript. The research questions, the experimental design, the interpretation of
results, and all claims in this paper are the authors' own; every reported number was
produced by running the released code and verified by the authors.

\bibliography{custom}

\clearpage

\appendix
\section{Model Architectures}
\label{app:model_details}

To ensure the universality of our \textbf{Alignment-Aggregation Divergence} hypothesis, we selected three Vision-Language Models (VLMs) that represent distinct design paradigms in the current landscape. Table \ref{tab:model_comparison} summarizes their architectural specifications.

\begin{table}[h]
\centering
\small
\begin{tabular}{l l c c}
\toprule
\textbf{Model} & \textbf{Backbone} & \textbf{Layers} & \textbf{Params} \\
\midrule
ColPali & \href{https://huggingface.co/google/paligemma-3b-mix-448}{PaliGemma-3B} & 18 & 3B \\
ColQwen2 & \href{https://huggingface.co/Qwen/Qwen2-VL-2B-Instruct}{Qwen2-VL-2B} & 28 & 2B \\
Jina v4 & \href{https://huggingface.co/Qwen/Qwen2.5-VL-3B-Instruct}{Qwen2.5-VL-3B} & 36 & 3B \\
\bottomrule
\end{tabular}
\caption{\textbf{Architectural Summary.} The selected models cover different LLM families (Gemma, Qwen, Llama-style) and vision encoding strategies. Note the progression from the fixed-resolution SigLIP encoder in PaliGemma to the dynamic-resolution capabilities inherent in the Qwen2-VL series.}
\label{tab:model_comparison}
\end{table}

\subsection{ColPali}
\textbf{ColPali} (\texttt{ViDoRe/colpali-v1.3}\footnote{\url{https://huggingface.co/ViDoRe/colpali-v1.3}}) represents the pioneering architecture for late-interaction visual retrieval, establishing the foundation for the {Visual Document Retrieval} (ViDoRe) benchmark. It is built upon the \textbf{PaliGemma-3B} backbone, which uniquely combines a \textbf{SigLIP-So400m} vision encoder with the Gemma-2B language model. Unlike traditional pipelines that rely on OCR to extract text, ColPali employs a {Visual Large Language Model} (VLLM) approach to generate multi-vector representations directly from document images. This allows it to effectively index complex visual elements—such as figures, charts, and tables—thereby significantly outperforming standard dense retrieval methods on visually rich documents.

\subsection{ColQwen2}
\textbf{ColQwen2} (\texttt{ViDoRe/colqwen2-v1.0}\footnote{\url{https://huggingface.co/ViDoRe/colqwen2-v1.0}}) is based on the advanced \textbf{Qwen2-VL-2B} architecture. This model introduces significant complexity and architectural improvements over ColPali, primarily through its support for \textbf{native dynamic resolution}. While ColPali typically resizes inputs to fixed square patches (often distorting document aspect ratios), ColQwen2 leverages the {Naive Dynamic Resolution} mechanism inherent to Qwen2-VL. This allows the model to process images of varying dimensions and aspect ratios without information loss, resulting in superior visual fidelity and more efficient visual token usage during the indexing of high-resolution PDFs.

\subsection{Jina Embeddings v4}
The \textbf{Jina Embeddings v4} model (\texttt{jinaai/jina-embeddings-v4}\footnote{\url{https://huggingface.co/jinaai/jina-embeddings-v4}}) represents the state-of-the-art application of late-interaction principles to the powerful \textbf{Qwen2.5-VL} architecture. By transitioning to the Qwen2.5 backbone, this iteration offers enhanced optical character recognition (OCR) capabilities and improved geometric reasoning for structured data. Furthermore, it incorporates Jina AI's signature \textbf{Matryoshka Representation Learning (MRL)}, enabling flexible embedding dimensions that allow users to trade off index vector size efficiency against retrieval precision. This model aims to unify multimodal retrieval by supporting extended context windows and delivering high-performance indexing for both textual and visual-heavy datasets.

\section{Baseline Implementation Details}
\label{app:baseline_details}

We provide the formal definitions for the baseline pruning methods used in our comparative analysis. Let $E \in \mathbb{R}^{N \times d}$ denote the sequence of $N$ visual patch embeddings. We select a subset of size $K$ based on the following importance scoring functions $S(j)$ for the $j$-th patch.

\paragraph{1. Random.}
This method assumes visual information is holographically distributed. The selection is performed via uniform pruning without replacement:
\begin{equation}
    S_{random}(j) \sim \mathcal{U}(0, 1)
\end{equation}
To mitigate the impact of randomness, we conducted five independent runs initialized with distinct random seeds and reported the average performance metrics.

\paragraph{2. EOS-Attention.}
This method assumes that patches attended to during text generation are most relevant. We define the score as the cross-attention weight from the final token (representing \texttt{\texttt{[EOS]}}) in the last layer $L_{last}$:
\begin{equation}
    S_{eos}(j) = \frac{1}{H} \sum_{h=1}^{H} A^{(L_{last}, h)}_{eos, j}
\end{equation}

\paragraph{3. Semantic Clustering (Post-Projector).}
This method assumes that visual redundancy can be reduced by grouping embeddings based on their representation similarity. Following the architectural insights from Light-ColPali~\cite{ma-etal-2025-towards-storage}, we specifically perform this operation at the \textbf{Post-Projector} stage—immediately after the Vision-LLM's final linear projection layer. 

The empirical study indicates that clustering is significantly more effective in this low-dimensional output space (e.g., 128 dimensions) compared to high-dimensional intermediate representations, as it enables more targeted feature aggregation with minimal information loss. We apply K-Means clustering to the set of {projected} embeddings $E = \{v_1, \dots, v_N\}$ to partition them into $K$ disjoint sets $\{C_1, \dots, C_K\}$. The objective is to minimize the within-cluster sum of squares (WCSS):
\begin{equation}
    \min_{\{\mu_1, \dots, \mu_K\}} \sum_{k=1}^{K} \sum_{v_j \in C_k} \| v_j - \mu_k \|^2
\end{equation}
where $\mu_k$ is the centroid of cluster $C_k$. The pruned representation consists of these $K$ centroids, effectively merging redundant visual features into a compact, representative set ready for indexing.

\paragraph{4. Adaptive EOS attention.}
While the standard EOS-Attention method applies a fixed selection ratio (Top-K) across all documents, this baseline adopts a document-aware adaptive thresholding strategy inspired by DocPruner \cite{yan2025docprunerstorageefficientframeworkmultivector}. This method postulates that the information density varies across documents, and thus the number of retained tokens should be dynamic.

For a document $d$, we compute the mean $\mu_d$ and standard deviation $\sigma_d$ of its patch importance scores $S_{eos}$. A patch $j$ is retained if its score exceeds a statistical threshold:
\begin{equation}
    S_{eos}(j) > \mu_d + k \cdot \sigma_d
\end{equation}
where $k$ is an adaptation factor controlling the aggressiveness of pruning.

\textbf{Fairness Calibration.} To ensure a rigorous comparison with fixed-ratio methods (like SAP) at a specific target retention ratio $\gamma$ (e.g., 10\%), we do not arbitrarily select $k$. Instead, we employ a calibration process. We extract the EOS attention scores from a held-out calibration set of 128 randomly sampled documents. We convert these scores into Z-scores $z_{ij} = (S_{eos}(j) - \mu_i) / \sigma_i$ and compute the global empirical quantile:
\begin{equation}
    k = \text{Quantile}(\{z_{ij}\}_{\text{calib}}, 1 - \gamma)
\end{equation}
This ensures that the {global average} retention rate of this adaptive baseline strictly aligns with the target $\gamma$, isolating the impact of the selection strategy (Adaptive vs. Fixed) from the index vector size budget.

\paragraph{Native operating point.} The quantile calibration above exists to
budget-match DocPruner to a target $\gamma$. To confirm it does not
misrepresent DocPruner's intended behavior, we additionally evaluate its
native document-adaptive thresholding with no calibration, retaining every
patch whose per-document $z$-score exceeds $\lambda$, for
$\lambda \in \{0, 0.5, 1.0, 1.5\}$. We record the resulting effective average
retention ratio and evaluate fixed-ratio SAP at that same per-dataset budget.
Table~\ref{tab:docpruner_native} reports the least aggressive point
($\lambda=0$); SAP leads on every backbone and both benchmarks, and the ordering
is unchanged at all four $\lambda$ values. The two operating modes are also
close to each other once budget-matched: on ColPali at an effective
$\gamma \approx 0.05$, the native ($\lambda=0.5$) and quantile-calibrated
($\gamma=0.05$) variants score 69.9/56.4 and 70.5/55.4 NDCG@5 retention on
v1/v2, within roughly one point.

\begin{table}[h]
\centering
\small
\setlength{\tabcolsep}{3.5pt}
\caption{\textbf{DocPruner at its native operating point} ($\lambda=0$, no
quantile calibration). NDCG@5 retention (\%) on ViDoRe v1 / v2. ``Eff.\
$\gamma$'' is the retention ratio the native rule actually produces on v1
(v2 values are within $0.02$: $0.148$, $0.245$, $0.330$). SAP is budget-matched
to that effective ratio per dataset.}
\label{tab:docpruner_native}
\begin{tabular}{l c c c}
\toprule
\textbf{Backbone} & \textbf{Eff.\ $\gamma$} & \textbf{DocPruner} & \textbf{SAP} \\
\midrule
ColPali  & 0.146 & 85.0 / 76.3 & \textbf{96.0} / \textbf{91.5} \\
ColQwen2 & 0.236 & 89.2 / 90.0 & \textbf{97.4} / \textbf{92.9} \\
Jina v4  & 0.317 & 97.2 / 95.4 & \textbf{99.4} / \textbf{96.3} \\
\bottomrule
\end{tabular}
\end{table}

\section{Dataset Details}
\label{app:dataset_details}
Table \ref{tab:dataset_specs} details the composition of the ViDoRe v1 and v2 benchmarks used in our comprehensive evaluation.

\begin{table}[h]
\centering
\small
\begin{tabular}{l l l}
\toprule
\textbf{Benchmark} & \textbf{Subset Name} & \textbf{Primary Domain} \\
\midrule
\multirow{8}{*}{\textbf{ViDoRe v1}} 
& ArxivQA & Academic / STEM \\
& DocVQA & General Document \\
& InfoVQA & Infographics / Layout \\
& Shift Project & Environmental Reports \\
& Artificial Intelligence & Technical Reports \\
& Energy & Industry Reports \\
& Government Reports & Policy / Legal \\
& Healthcare Industry & Medical / Business \\
\midrule
\multirow{4}{*}{\textbf{ViDoRe v2}} 
& MIT Biomedical (Multi) & Medical / Research \\
& Economics Macro (Multi) & Finance / Policy \\
& ESG Restaurant (Multi) & Business / Tables \\
& ESG Restaurant (Human) & Business / Tables \\
\bottomrule
\end{tabular}
\caption{\textbf{Dataset Specifications.} Overview of domains covered in the ViDoRe benchmark suite.}
\label{tab:dataset_specs}
\end{table}

\section{SR-Guided Window Selection: Formal Algorithm}
\label{app:osr_algorithm}

We provide formal pseudocode for the SR-guided window selection procedure described in Section~\ref{sec:osr_window}.

\begin{algorithm}[t]
\caption{SR-Guided Window Selection}
\label{alg:osr_window}
\begin{algorithmic}[1]
\Require Backbone $\mathcal{M}$ with $L_{total}$ layers; calibration set
         $\mathcal{C} = \{(I_n, Q_n)\}_{n=1}^N$; retention ratio $\gamma$;
         relative window width $\rho$.
\Ensure  Window $(\alpha, \beta)$.
\State $r_l \gets 0$ for all $l \in \{0, \dots, L_{total}-1\}$
\For{$(I, Q) \in \mathcal{C}$}
    \State Forward $I$ through $\mathcal{M}$ to obtain per-layer attention
           tensors $\{A^{(l)}\}$ and full visual embeddings $E_I$.
    \State Encode query: $E_Q \gets \mathcal{M}_{\text{txt}}(Q)$.
    \State $\text{full} \gets \mathrm{MaxSim}(E_Q, E_I)$
    \For{$l \in \{0, \dots, L_{total}-1\}$}
        \State $c \gets S^{(l)}$ from $A^{(l)}$ \Comment{Eq.~3}
        \State $\hat{E}_I \gets \text{top-}\lceil\gamma|E_I|\rceil(E_I, c)$
        \State $r_l \gets r_l + \mathrm{MaxSim}(E_Q, \hat{E}_I) / \text{full}$ \Comment{SR, Eq.~\ref{eq:sr}}
    \EndFor
\EndFor
\State $\mathcal{R}(l) \gets r_l / N$ for all $l$
\State $m \gets \mathrm{median}\{\mathcal{R}(l)\}_{l=0}^{L_{total}-1}$
\State $l^* \gets$ smallest index s.t.\ $\mathcal{R}(l) < m$ for all $l \in [l^*, L_{total}-1]$
\State $k \gets \lceil \rho \cdot L_{total} \rceil$
\State $\beta \gets l^* / L_{total}$;\quad $\alpha \gets (l^* - k) / L_{total}$
\State \Return $(\alpha, \beta)$
\end{algorithmic}
\end{algorithm}

\paragraph{Calibration cost.} On a V100-32GB GPU, the procedure completes within $\sim$30~min for ColPali (18 layers) and within $\sim$2~h for Jina~v4 (36 layers) with $N=500$ pairs. Since the window selection runs once per backbone, offline, this cost is amortized across all subsequent queries.

\section{Per-Architecture Window Instantiation}
\label{app:layer_instantiation}

The SR-guided window selection procedure  is run once per backbone on 500 (image, query) pairs uniformly sampled from the ColPali training corpus (\texttt{vidore/colpali\_train\_set}), which is disjoint from all ViDoRe evaluation splits. Table~\ref{tab:layer_instantiation} reports the resulting per-architecture windows together with a head-to-head comparison against the brute-force best window from Table~\ref{tab:window_ablation}. Two reference points matter here and the submitted version conflated them. Against the brute-force best window on the \emph{mean} of ViDoRe v1 and v2 --- the criterion the procedure targets --- the SR-guided window is itself optimal for all three backbones, so the mean gap is $0.0$. Against the best window on each benchmark \emph{separately}, ColPali trails the 50--70\% window by $0.3$ NDCG@5 on v1 and Jina~v4 trails it by $0.2$ on v2, while ColQwen2 matches on both. Table~\ref{tab:layer_instantiation} reports both quantities.. The alignment region (suffix layers with $\mathcal{R}(l) < m$) confirms an architecture-invariant pattern: 2--4 final layers dominated by retrieval-objective alignment.

\begin{table}[h]
\centering
\small
\setlength{\tabcolsep}{3pt}
\caption{\textbf{SR-guided window selection per backbone} ($\gamma=0.10$, 500-pair calibration set). The procedure  automatically determines the window position; width is fixed at $k = \lceil 0.2\,L \rceil$. ``Gap (mean)'' is the NDCG@5 retention gap between the SR-guided window and the brute-force best window on the v1+v2 mean; ``Gap (worst)'' is the largest gap against the best window on either benchmark taken separately. Both are computed from Table~\ref{tab:window_ablation}.}
\label{tab:layer_instantiation}
\begin{tabular}{l c c c c c c}
\toprule
\textbf{Backbone} & $L$ & $l^*$ & \textbf{Layers} & \textbf{Snap} & \textbf{Gap (mean)} & \textbf{Gap (worst)} \\
\midrule
ColPali  & 18 & 15 & 11--14 & 60--80\% & $0.0$ & $-0.3$ \\
ColQwen2 & 28 & 24 & 18--23 & 60--80\% & $0.0$ & $\phantom{-}0.0$ \\
Jina v4  & 36 & 34 & 26--33 & 70--90\% & $0.0$ & $-0.2$ \\
\bottomrule
\end{tabular}
\end{table}

\section{Three Typical Compression Ratios: Detailed Case Study}
\label{app:three_gamma}

To complement the continuous $\gamma$ sweep in Figure~\ref{fig:retention_curve}, Table~\ref{tab:global_results} reports the three operational regimes most commonly cited in prior work~\cite{ma-etal-2025-towards-storage,yan2025docprunerstorageefficientframeworkmultivector}: moderate compression ($\gamma=0.20$, $5\times$), aggressive ($\gamma=0.10$, $10\times$), and extreme ($\gamma=0.05$, $20\times$). For each (backbone, benchmark, method, $\gamma$) we report Score Retention (Eq.~\ref{eq:sr}), absolute NDCG@5, and the NDCG@5 retention percentage relative to the full-model upper bound.

\begin{table*}[t]
\centering
\footnotesize
\setlength{\tabcolsep}{2.8pt}
\resizebox{0.95\textwidth}{!}{
\begin{tabular}{l l c ccc ccc ccc}
\toprule
\multirow{2}{*}{\textbf{Model}} & \multirow{2}{*}{\textbf{Method}} & \textbf{Upper Bound} & \multicolumn{3}{c}{$\boldsymbol{\gamma=0.20}$} & \multicolumn{3}{c}{$\boldsymbol{\gamma=0.10}$} & \multicolumn{3}{c}{$\boldsymbol{\gamma=0.05}$} \\
\cmidrule(lr){3-3} \cmidrule(lr){4-6} \cmidrule(lr){7-9} \cmidrule(lr){10-12}
& & \textbf{Full NDCG@5} & \textbf{S.Ret} & \textbf{NDCG@5} & \textbf{\%} & \textbf{S.Ret} & \textbf{NDCG@5} & \textbf{\%} & \textbf{S.Ret} & \textbf{NDCG@5} & \textbf{\%} \\
\midrule
\multicolumn{12}{c}{\cellcolor{gray!10}\textbf{Benchmark: ViDoRe v1 (Avg. across 10 datasets)}} \\
\midrule
\multirow{4}{*}{\textbf{ColPali}} & EOS-Adaptive & \multirow{4}{*}{0.85} & 0.86 & 0.76 & 89.02 & 0.79 & 0.70 & 81.16 & 0.72 & 0.63 & 73.26 \\
 & Random &  & 0.91 & 0.81 & 94.61 & 0.85 & 0.77 & 89.82 & 0.78 & 0.71 & 82.92 \\
 & Cluster &  & 0.89 & 0.82 & 96.53 & 0.80 & 0.77 & 90.42 & 0.69 & 0.67 & 78.22 \\
 & \textbf{SAP (Ours)} &  & \textbf{0.94} & \textbf{0.83} & \textbf{97.48} & \textbf{0.89} & \textbf{0.80} & \textbf{93.88} & \textbf{0.81} & \textbf{0.74} & \textbf{86.34} \\
\cmidrule{1-12}
\multirow{4}{*}{\textbf{ColQwen2}} & EOS-Adaptive & \multirow{4}{*}{0.88} & 0.84 & 0.77 & 87.90 & 0.75 & 0.70 & 79.54 & 0.66 & 0.63 & 71.18 \\
 & Random &  & 0.86 & 0.83 & 94.10 & 0.78 & 0.78 & 87.83 & 0.68 & 0.70 & 78.65 \\
 & Cluster &  & 0.82 & 0.84 & 95.97 & 0.70 & 0.79 & 90.14 & 0.59 & 0.73 & 82.66 \\
 & \textbf{SAP (Ours)} &  & \textbf{0.92} & \textbf{0.85} & \textbf{97.08} & \textbf{0.86} & \textbf{0.82} & \textbf{93.23} & \textbf{0.78} & \textbf{0.75} & \textbf{85.56} \\
\cmidrule{1-12}
\multirow{4}{*}{\textbf{Jina Embeddings v4}} & EOS-Adaptive & \multirow{4}{*}{0.90} & 0.87 & 0.85 & 95.07 & 0.79 & 0.79 & 87.61 & 0.71 & 0.67 & 73.36 \\
 & Random &  & 0.89 & 0.86 & 95.21 & 0.82 & 0.82 & 90.42 & 0.74 & 0.75 & 82.40 \\
 & Cluster &  & 0.83 & 0.86 & 95.97 & 0.73 & 0.81 & 90.37 & 0.62 & 0.75 & 83.23 \\
 & \textbf{SAP (Ours)} &  & \textbf{0.91} & \textbf{0.88} & \textbf{98.42} & \textbf{0.85} & \textbf{0.86} & \textbf{95.57} & \textbf{0.78} & \textbf{0.81} & \textbf{89.55} \\
\midrule
\multicolumn{12}{c}{\cellcolor{gray!10}\textbf{Benchmark: ViDoRe v2 (Avg. across 4 datasets)}} \\
\midrule
\multirow{4}{*}{\textbf{ColPali}} & EOS-Adaptive & \multirow{4}{*}{0.56} & 0.87 & 0.45 & 80.93 & 0.80 & 0.40 & 71.85 & 0.72 & 0.33 & 59.52 \\
 & Random &  & 0.90 & 0.49 & 87.72 & 0.84 & 0.44 & 78.54 & 0.77 & 0.38 & 69.42 \\
 & Cluster &  & 0.88 & 0.46 & 83.40 & 0.79 & 0.38 & 69.00 & 0.67 & 0.30 & 54.37 \\
 & \textbf{SAP (Ours)} &  & \textbf{0.94} & \textbf{0.52} & \textbf{93.44} & \textbf{0.89} & \textbf{0.50} & \textbf{90.38} & \textbf{0.81} & \textbf{0.42} & \textbf{76.39} \\
\cmidrule{1-12}
\multirow{4}{*}{\textbf{ColQwen2}} & EOS-Adaptive & \multirow{4}{*}{0.54} & 0.84 & 0.48 & 89.26 & 0.76 & 0.43 & 79.92 & 0.67 & 0.35 & 64.33 \\
 & Random &  & 0.87 & 0.46 & 86.04 & 0.78 & 0.41 & 76.12 & 0.69 & 0.35 & 65.15 \\
 & Cluster &  & 0.80 & 0.45 & 83.56 & 0.70 & 0.41 & 76.25 & 0.59 & 0.33 & 61.68 \\
 & \textbf{SAP (Ours)} &  & \textbf{0.91} & \textbf{0.50} & \textbf{91.91} & \textbf{0.85} & \textbf{0.48} & \textbf{88.07} & \textbf{0.77} & \textbf{0.43} & \textbf{78.96} \\
\cmidrule{1-12}
\multirow{4}{*}{\textbf{Jina Embeddings v4}} & EOS-Adaptive & \multirow{4}{*}{0.58} & 0.90 & 0.53 & 90.82 & 0.82 & 0.49 & 83.73 & 0.73 & 0.38 & 65.31 \\
 & Random &  & 0.89 & 0.51 & 87.78 & 0.83 & 0.47 & 80.44 & 0.75 & 0.39 & 67.51 \\
 & Cluster &  & 0.83 & 0.47 & 80.74 & 0.73 & 0.40 & 67.99 & 0.63 & 0.33 & 57.14 \\
 & \textbf{SAP (Ours)} &  & \textbf{0.92} & \textbf{0.56} & \textbf{95.78} & \textbf{0.87} & \textbf{0.51} & \textbf{88.52} & \textbf{0.81} & \textbf{0.44} & \textbf{76.51} \\
\bottomrule
\end{tabular}
}
\caption{\textbf{Three typical compression ratios.} NDCG@5 retention (\%) for training-free compression methods at $\gamma \in \{0.20, 0.10, 0.05\}$ across three VLM backbones and two benchmark suites. \textbf{S.Ret} (Score Retention, Eq.~\ref{eq:sr}) diagnoses per-pair MaxSim fidelity; the \textbf{\%} column reports $(\text{NDCG@5}_{\text{pruned}} / \text{NDCG@5}_{\text{full}}) \times 100$. SAP uses our SR-guided window (ColPali \& ColQwen2: 60--80\%, Jina v4: 70--90\%; see Section~\ref{sec:osr_window}). Best per column in \textbf{bold}.}
\label{tab:global_results}
\end{table*}

\paragraph{Performance Consistency at $\gamma=0.10$.}
At $\gamma=0.10$, the standard $10\times$ target, SAP retains 93--96\% of NDCG@5 on ViDoRe v1 and 88--90\% on ViDoRe v2. At $\gamma=0.05$ on ViDoRe v2, where baselines drop to 54--69\%, SAP maintains 76--79\%, leading the strongest baseline on each backbone by $7$--$14$ points and the weakest by up to $22$, closing the high-compression gap previously cited as a barrier to training-free deployment.

\paragraph{Universality at the same $\gamma$.}
The same three $\gamma$ values yield qualitatively identical method rankings across backbones spanning 18 to 36 layers: SAP $>$ Cluster/Random $>$ EOS-Adaptive at $\gamma=0.20$, and the gap widens as $\gamma$ shrinks. No per-architecture tuning of SAP is performed.

\paragraph{Cross-Cutoff Stability.}
To verify the $\gamma=0.10$ numbers are not artifacts of NDCG@5 specifically, Appendix~\ref{app:ndcg_cutoffs} reports NDCG@$\{1, 5, 10, 50, 100\}$ retention for SAP at all three typical $\gamma$ values. The spread across cutoffs is below $2.5\%$ at $\gamma=0.20$ and below $8\%$ at $\gamma=0.05$ on both benchmarks, confirming SAP exhibits no cutoff-sensitive degradation. Full per-dataset numbers across all 14 ViDoRe datasets are provided next in Appendix~\ref{app:results_v1}.

\clearpage

\section{Detailed Evaluation Results}
\label{app:results_v1}

Extending the aggregated case study in Appendix~\ref{app:three_gamma}, we report the full per-dataset NDCG@5 retention for SAP at the SR-guided window against the three training-free baselines on each of the 14 ViDoRe datasets. Tables~\ref{tab:v1_colpali}--\ref{tab:v1_jina} cover ViDoRe v1 for ColPali, ColQwen2, and Jina v4 respectively; Tables~\ref{tab:v2_colpali}--\ref{tab:v2_jina} cover ViDoRe v2 for the same three backbones. SAP's relative advantage varies across datasets; the variance is largely explained by the per-architecture optimal window position revealed by our SR-guided procedure .

\clearpage
\begin{table*}[p]
\centering
\scriptsize
\setlength{\tabcolsep}{1.8pt}
\renewcommand{\arraystretch}{1.1}
\resizebox{0.9\textwidth}{!}{
\begin{tabular}{l l c ccc ccc ccc}
\toprule
\multirow{2}{*}{\textbf{Dataset}} & \multirow{2}{*}{\textbf{Method}} & \textbf{Upper Bound} & \multicolumn{3}{c}{$\boldsymbol{\gamma=0.20}$} & \multicolumn{3}{c}{$\boldsymbol{\gamma=0.10}$} & \multicolumn{3}{c}{$\boldsymbol{\gamma=0.05}$} \\
\cmidrule(lr){3-3} \cmidrule(lr){4-6} \cmidrule(lr){7-9} \cmidrule(lr){10-12}
& & \textbf{Full NDCG@5} & \textbf{S.Ret} & \textbf{NDCG@5} & \textbf{\%} & \textbf{S.Ret} & \textbf{NDCG@5} & \textbf{\%} & \textbf{S.Ret} & \textbf{NDCG@5} & \textbf{\%} \\
\midrule
\multirow{4}{*}{ArxivQA} & EOS-Adaptive & \multirow{4}{*}{0.82} & 0.91 & 0.77 & 93.82 & 0.85 & 0.73 & 89.08 & 0.78 & 0.68 & 83.67 \\
 & Random &  & 0.93 & 0.80 & 98.04 & 0.89 & 0.78 & 95.90 & 0.83 & 0.75 & 91.29 \\
 & Cluster &  & 0.92 & 0.81 & 99.07 & 0.87 & 0.80 & 98.07 & \textbf{0.80} & \textbf{0.79} & \textbf{96.25} \\
 & \textbf{SAP (Ours)} &  & \textbf{0.95} & \textbf{0.82} & \textbf{99.75} & \textbf{0.90} & \textbf{0.81} & \textbf{98.87} & 0.84 & 0.77 & 94.25 \\
\cmidrule{1-12}
\multirow{4}{*}{DocVQA} & EOS-Adaptive & \multirow{4}{*}{0.59} & 0.91 & 0.50 & 84.18 & 0.86 & 0.43 & 72.33 & 0.81 & 0.35 & 59.41 \\
 & Random &  & 0.94 & 0.54 & 91.65 & 0.90 & 0.50 & 85.03 & 0.84 & 0.43 & 73.58 \\
 & Cluster &  & 0.92 & 0.55 & 93.44 & 0.86 & 0.51 & 86.09 & 0.77 & 0.45 & 76.05 \\
 & \textbf{SAP (Ours)} &  & \textbf{0.97} & \textbf{0.57} & \textbf{96.51} & \textbf{0.93} & \textbf{0.52} & \textbf{88.71} & \textbf{0.86} & \textbf{0.46} & \textbf{78.82} \\
\cmidrule{1-12}
\multirow{4}{*}{InfoVQA} & EOS-Adaptive & \multirow{4}{*}{0.85} & 0.86 & 0.77 & 89.92 & 0.79 & 0.72 & 84.87 & 0.73 & 0.68 & 79.50 \\
 & Random &  & 0.91 & 0.82 & 96.37 & 0.86 & 0.79 & 93.16 & \textbf{0.79} & \textbf{0.76} & \textbf{89.42} \\
 & Cluster &  & \textbf{0.90} & \textbf{0.84} & \textbf{98.04} & 0.82 & 0.81 & 94.64 & 0.71 & 0.74 & 87.10 \\
 & \textbf{SAP (Ours)} &  & 0.94 & 0.83 & 97.63 & \textbf{0.88} & \textbf{0.81} & \textbf{94.96} & 0.80 & 0.75 & 87.84 \\
\cmidrule{1-12}
\multirow{4}{*}{Shift Project} & EOS-Adaptive & \multirow{4}{*}{0.77} & 0.83 & 0.60 & 77.60 & 0.75 & 0.48 & 62.41 & 0.66 & 0.40 & 51.41 \\
 & Random &  & 0.90 & 0.68 & 88.04 & 0.84 & 0.61 & 78.99 & 0.76 & 0.51 & 65.23 \\
 & Cluster &  & 0.85 & 0.68 & 87.25 & 0.74 & 0.53 & 68.78 & 0.60 & 0.30 & 38.82 \\
 & \textbf{SAP (Ours)} &  & \textbf{0.94} & \textbf{0.74} & \textbf{95.29} & \textbf{0.88} & \textbf{0.66} & \textbf{85.45} & \textbf{0.81} & \textbf{0.57} & \textbf{73.24} \\
\cmidrule{1-12}
\multirow{4}{*}{SynDocQA-AI} & EOS-Adaptive & \multirow{4}{*}{0.97} & 0.87 & 0.94 & 96.67 & 0.80 & 0.88 & 90.58 & 0.74 & 0.86 & 88.11 \\
 & Random &  & 0.90 & 0.95 & 97.96 & 0.84 & 0.92 & 95.04 & 0.76 & 0.87 & 89.56 \\
 & Cluster &  & \textbf{0.88} & \textbf{0.96} & \textbf{99.04} & 0.78 & 0.92 & 94.37 & 0.67 & 0.82 & 84.51 \\
 & \textbf{SAP (Ours)} &  & 0.94 & 0.95 & 98.08 & \textbf{0.89} & \textbf{0.93} & \textbf{95.77} & \textbf{0.81} & \textbf{0.90} & \textbf{92.06} \\
\cmidrule{1-12}
\multirow{4}{*}{SynDocQA-Energy} & EOS-Adaptive & \multirow{4}{*}{0.96} & 0.84 & 0.87 & 90.69 & 0.76 & 0.77 & 79.94 & 0.69 & 0.75 & 78.23 \\
 & Random &  & 0.90 & 0.93 & 96.87 & 0.84 & 0.90 & 93.50 & 0.76 & 0.85 & 88.23 \\
 & Cluster &  & \textbf{0.89} & \textbf{0.95} & \textbf{98.76} & 0.80 & 0.92 & 95.66 & 0.67 & 0.80 & 83.65 \\
 & \textbf{SAP (Ours)} &  & 0.94 & 0.93 & 97.15 & \textbf{0.89} & \textbf{0.92} & \textbf{96.30} & \textbf{0.80} & \textbf{0.85} & \textbf{89.09} \\
\cmidrule{1-12}
\multirow{4}{*}{SynDocQA-Gov} & EOS-Adaptive & \multirow{4}{*}{0.96} & 0.85 & 0.90 & 92.92 & 0.79 & 0.86 & 88.97 & 0.70 & 0.74 & 76.95 \\
 & Random &  & 0.91 & 0.92 & 95.05 & 0.84 & 0.88 & 91.38 & 0.77 & 0.84 & 86.68 \\
 & Cluster &  & 0.88 & 0.94 & 97.42 & 0.78 & 0.90 & 93.57 & 0.65 & 0.78 & 81.13 \\
 & \textbf{SAP (Ours)} &  & \textbf{0.95} & \textbf{0.95} & \textbf{98.60} & \textbf{0.90} & \textbf{0.94} & \textbf{97.36} & \textbf{0.80} & \textbf{0.86} & \textbf{89.21} \\
\cmidrule{1-12}
\multirow{4}{*}{SynDocQA-Health} & EOS-Adaptive & \multirow{4}{*}{0.97} & 0.87 & 0.93 & 96.35 & 0.81 & 0.89 & 91.89 & 0.74 & 0.80 & 82.64 \\
 & Random &  & 0.91 & 0.94 & 97.57 & 0.84 & 0.91 & 94.16 & 0.78 & 0.87 & 90.04 \\
 & Cluster &  & 0.88 & 0.95 & 98.37 & 0.77 & 0.88 & 91.05 & 0.65 & 0.70 & 72.73 \\
 & \textbf{SAP (Ours)} &  & \textbf{0.95} & \textbf{0.95} & \textbf{98.74} & \textbf{0.90} & \textbf{0.95} & \textbf{98.53} & \textbf{0.82} & \textbf{0.90} & \textbf{93.35} \\
\cmidrule{1-12}
\multirow{4}{*}{TabFQuAD} & EOS-Adaptive & \multirow{4}{*}{0.87} & 0.82 & 0.79 & 90.95 & 0.75 & 0.76 & 86.90 & 0.67 & 0.70 & 80.39 \\
 & Random &  & 0.91 & 0.85 & 97.92 & 0.85 & 0.83 & 95.40 & 0.77 & 0.79 & 91.13 \\
 & Cluster &  & \textbf{0.91} & \textbf{0.88} & \textbf{100.53} & \textbf{0.85} & \textbf{0.87} & \textbf{100.11} & \textbf{0.77} & \textbf{0.86} & \textbf{98.70} \\
 & \textbf{SAP (Ours)} &  & 0.94 & 0.86 & 99.13 & 0.89 & 0.85 & 97.10 & 0.81 & 0.82 & 94.09 \\
\cmidrule{1-12}
\multirow{4}{*}{TAT-DQA} & EOS-Adaptive & \multirow{4}{*}{0.71} & 0.85 & 0.55 & 77.11 & 0.79 & 0.46 & 64.66 & 0.73 & 0.37 & 52.34 \\
 & Random &  & 0.89 & 0.62 & 86.58 & 0.81 & 0.54 & 75.70 & 0.73 & 0.46 & 64.04 \\
 & Cluster &  & 0.86 & 0.67 & 93.35 & 0.76 & 0.58 & 81.85 & 0.64 & 0.45 & 63.24 \\
 & \textbf{SAP (Ours)} &  & \textbf{0.94} & \textbf{0.67} & \textbf{93.91} & \textbf{0.88} & \textbf{0.61} & \textbf{85.73} & \textbf{0.79} & \textbf{0.51} & \textbf{71.41} \\
\bottomrule
\end{tabular}
}
\caption{\textbf{Detailed Performance: ColPali on ViDoRe v1.} Per-dataset NDCG@5 retention with SAP using the SR-guided window . Best per column in \textbf{bold}.}
\label{tab:v1_colpali}
\end{table*}
\clearpage

\begin{table*}[p]
\centering
\scriptsize
\setlength{\tabcolsep}{1.8pt}
\renewcommand{\arraystretch}{1.1}
\resizebox{0.9\textwidth}{!}{
\begin{tabular}{l l c ccc ccc ccc}
\toprule
\multirow{2}{*}{\textbf{Dataset}} & \multirow{2}{*}{\textbf{Method}} & \textbf{Upper Bound} & \multicolumn{3}{c}{$\boldsymbol{\gamma=0.20}$} & \multicolumn{3}{c}{$\boldsymbol{\gamma=0.10}$} & \multicolumn{3}{c}{$\boldsymbol{\gamma=0.05}$} \\
\cmidrule(lr){3-3} \cmidrule(lr){4-6} \cmidrule(lr){7-9} \cmidrule(lr){10-12}
& & \textbf{Full NDCG@5} & \textbf{S.Ret} & \textbf{NDCG@5} & \textbf{\%} & \textbf{S.Ret} & \textbf{NDCG@5} & \textbf{\%} & \textbf{S.Ret} & \textbf{NDCG@5} & \textbf{\%} \\
\midrule
\multirow{4}{*}{ArxivQA} & EOS-Adaptive & \multirow{4}{*}{0.86} & 0.85 & 0.76 & 88.38 & 0.76 & 0.71 & 82.59 & 0.66 & 0.63 & 73.50 \\
 & Random &  & 0.91 & 0.82 & 95.74 & 0.84 & 0.79 & 92.22 & 0.76 & 0.75 & 86.82 \\
 & Cluster &  & \textbf{0.87} & \textbf{0.84} & \textbf{97.16} & \textbf{0.79} & \textbf{0.82} & \textbf{95.65} & \textbf{0.69} & \textbf{0.79} & \textbf{92.03} \\
 & \textbf{SAP (Ours)} &  & 0.93 & 0.82 & 95.48 & 0.87 & 0.79 & 91.62 & 0.78 & 0.74 & 85.42 \\
\cmidrule{1-12}
\multirow{4}{*}{DocVQA} & EOS-Adaptive & \multirow{4}{*}{0.59} & 0.87 & 0.49 & 83.27 & 0.80 & 0.40 & 69.00 & 0.71 & 0.30 & 51.04 \\
 & Random &  & 0.86 & 0.52 & 89.04 & 0.77 & 0.46 & 77.75 & 0.66 & 0.37 & 62.83 \\
 & Cluster &  & 0.82 & 0.56 & 94.74 & 0.71 & 0.51 & 87.41 & 0.59 & 0.44 & 75.37 \\
 & \textbf{SAP (Ours)} &  & \textbf{0.93} & \textbf{0.56} & \textbf{95.62} & \textbf{0.88} & \textbf{0.52} & \textbf{88.45} & \textbf{0.81} & \textbf{0.47} & \textbf{79.90} \\
\cmidrule{1-12}
\multirow{4}{*}{InfoVQA} & EOS-Adaptive & \multirow{4}{*}{0.91} & 0.83 & 0.82 & 90.90 & 0.74 & 0.76 & 84.26 & 0.65 & 0.72 & 78.91 \\
 & Random &  & 0.86 & 0.85 & 93.53 & 0.78 & 0.81 & 89.14 & 0.70 & 0.75 & 82.52 \\
 & Cluster &  & 0.81 & 0.87 & 95.89 & 0.69 & 0.81 & 89.70 & 0.56 & 0.72 & 79.25 \\
 & \textbf{SAP (Ours)} &  & \textbf{0.91} & \textbf{0.88} & \textbf{96.74} & \textbf{0.85} & \textbf{0.84} & \textbf{92.67} & \textbf{0.77} & \textbf{0.78} & \textbf{85.71} \\
\cmidrule{1-12}
\multirow{4}{*}{Shift Project} & EOS-Adaptive & \multirow{4}{*}{0.86} & 0.84 & 0.68 & 79.63 & 0.76 & 0.61 & 71.49 & 0.67 & 0.52 & 60.65 \\
 & Random &  & 0.85 & 0.77 & 90.03 & 0.76 & 0.66 & 77.74 & 0.66 & 0.53 & 61.49 \\
 & Cluster &  & 0.79 & 0.77 & 90.04 & 0.66 & 0.63 & 73.88 & 0.54 & 0.54 & 63.56 \\
 & \textbf{SAP (Ours)} &  & \textbf{0.91} & \textbf{0.83} & \textbf{97.59} & \textbf{0.85} & \textbf{0.81} & \textbf{95.04} & \textbf{0.76} & \textbf{0.69} & \textbf{80.36} \\
\cmidrule{1-12}
\multirow{4}{*}{SynDocQA-AI} & EOS-Adaptive & \multirow{4}{*}{0.98} & 0.82 & 0.86 & 87.29 & 0.74 & 0.83 & 84.51 & 0.64 & 0.79 & 80.91 \\
 & Random &  & 0.87 & 0.97 & 98.45 & 0.78 & 0.94 & 95.78 & 0.68 & 0.86 & 87.94 \\
 & Cluster &  & \textbf{0.81} & \textbf{0.98} & \textbf{99.69} & \textbf{0.70} & \textbf{0.96} & \textbf{97.27} & \textbf{0.58} & \textbf{0.88} & \textbf{89.57} \\
 & \textbf{SAP (Ours)} &  & 0.92 & 0.96 & 97.34 & 0.85 & 0.93 & 95.11 & 0.76 & 0.87 & 88.24 \\
\cmidrule{1-12}
\multirow{4}{*}{SynDocQA-Energy} & EOS-Adaptive & \multirow{4}{*}{0.96} & 0.82 & 0.84 & 87.61 & 0.72 & 0.70 & 72.79 & 0.62 & 0.66 & 69.48 \\
 & Random &  & 0.86 & 0.90 & 94.01 & 0.77 & 0.84 & 88.03 & 0.68 & 0.77 & 80.90 \\
 & Cluster &  & \textbf{0.82} & \textbf{0.92} & \textbf{96.26} & 0.70 & 0.88 & 92.40 & 0.58 & 0.83 & 86.58 \\
 & \textbf{SAP (Ours)} &  & 0.92 & 0.92 & 95.72 & \textbf{0.86} & \textbf{0.90} & \textbf{93.85} & \textbf{0.79} & \textbf{0.86} & \textbf{89.90} \\
\cmidrule{1-12}
\multirow{4}{*}{SynDocQA-Gov} & EOS-Adaptive & \multirow{4}{*}{0.94} & 0.84 & 0.91 & 96.20 & 0.74 & 0.84 & 88.93 & 0.64 & 0.76 & 80.86 \\
 & Random &  & 0.87 & 0.94 & 99.28 & 0.78 & 0.90 & 94.97 & 0.69 & 0.82 & 87.18 \\
 & Cluster &  & 0.80 & 0.91 & 96.55 & 0.69 & 0.88 & 93.37 & 0.58 & 0.85 & 89.74 \\
 & \textbf{SAP (Ours)} &  & \textbf{0.93} & \textbf{0.94} & \textbf{99.33} & \textbf{0.86} & \textbf{0.91} & \textbf{96.53} & \textbf{0.79} & \textbf{0.87} & \textbf{92.29} \\
\cmidrule{1-12}
\multirow{4}{*}{SynDocQA-Health} & EOS-Adaptive & \multirow{4}{*}{0.98} & 0.83 & 0.93 & 95.16 & 0.76 & 0.89 & 91.40 & 0.67 & 0.81 & 82.82 \\
 & Random &  & 0.86 & 0.96 & 97.88 & 0.78 & 0.93 & 95.21 & 0.70 & 0.88 & 89.69 \\
 & Cluster &  & \textbf{0.81} & \textbf{0.97} & \textbf{99.49} & 0.69 & 0.91 & 93.22 & 0.56 & 0.85 & 86.92 \\
 & \textbf{SAP (Ours)} &  & 0.93 & 0.97 & 98.81 & \textbf{0.87} & \textbf{0.97} & \textbf{98.81} & \textbf{0.79} & \textbf{0.91} & \textbf{93.15} \\
\cmidrule{1-12}
\multirow{4}{*}{TabFQuAD} & EOS-Adaptive & \multirow{4}{*}{0.88} & 0.87 & 0.81 & 92.38 & 0.78 & 0.74 & 84.01 & 0.69 & 0.68 & 77.87 \\
 & Random &  & 0.87 & 0.85 & 97.19 & 0.79 & 0.82 & 93.39 & 0.70 & 0.76 & 86.98 \\
 & Cluster &  & 0.85 & 0.86 & 98.41 & \textbf{0.75} & \textbf{0.85} & \textbf{96.65} & \textbf{0.64} & \textbf{0.83} & \textbf{94.09} \\
 & \textbf{SAP (Ours)} &  & \textbf{0.94} & \textbf{0.88} & \textbf{100.23} & 0.88 & 0.84 & 95.57 & 0.80 & 0.80 & 91.42 \\
\cmidrule{1-12}
\multirow{4}{*}{TAT-DQA} & EOS-Adaptive & \multirow{4}{*}{0.81} & 0.83 & 0.63 & 78.12 & 0.72 & 0.54 & 66.39 & 0.61 & 0.45 & 55.73 \\
 & Random &  & 0.82 & 0.70 & 85.89 & 0.72 & 0.60 & 74.11 & 0.61 & 0.49 & 60.12 \\
 & Cluster &  & 0.79 & 0.74 & 91.46 & 0.67 & 0.66 & 81.81 & \textbf{0.55} & \textbf{0.56} & \textbf{69.49} \\
 & \textbf{SAP (Ours)} &  & \textbf{0.92} & \textbf{0.76} & \textbf{93.95} & \textbf{0.84} & \textbf{0.69} & \textbf{84.71} & 0.72 & 0.56 & 69.19 \\
\bottomrule
\end{tabular}
}
\caption{\textbf{Detailed Performance: ColQwen2 on ViDoRe v1.} Per-dataset NDCG@5 retention with SAP using the SR-guided window . Best per column in \textbf{bold}.}
\label{tab:v1_colqwen}
\end{table*}
\clearpage

\begin{table*}[p]
\centering
\scriptsize
\setlength{\tabcolsep}{1.8pt}
\renewcommand{\arraystretch}{1.1}
\resizebox{0.9\textwidth}{!}{
\begin{tabular}{l l c ccc ccc ccc}
\toprule
\multirow{2}{*}{\textbf{Dataset}} & \multirow{2}{*}{\textbf{Method}} & \textbf{Upper Bound} & \multicolumn{3}{c}{$\boldsymbol{\gamma=0.20}$} & \multicolumn{3}{c}{$\boldsymbol{\gamma=0.10}$} & \multicolumn{3}{c}{$\boldsymbol{\gamma=0.05}$} \\
\cmidrule(lr){3-3} \cmidrule(lr){4-6} \cmidrule(lr){7-9} \cmidrule(lr){10-12}
& & \textbf{Full NDCG@5} & \textbf{S.Ret} & \textbf{NDCG@5} & \textbf{\%} & \textbf{S.Ret} & \textbf{NDCG@5} & \textbf{\%} & \textbf{S.Ret} & \textbf{NDCG@5} & \textbf{\%} \\
\midrule
\multirow{4}{*}{ArxivQA} & EOS-Adaptive & \multirow{4}{*}{0.88} & 0.90 & 0.84 & 94.93 & 0.82 & 0.77 & 86.81 & 0.73 & 0.64 & 72.19 \\
 & Random &  & 0.92 & 0.86 & 96.85 & 0.86 & 0.82 & 93.14 & 0.79 & 0.77 & 87.33 \\
 & Cluster &  & \textbf{0.88} & \textbf{0.87} & \textbf{98.80} & \textbf{0.81} & \textbf{0.85} & \textbf{95.83} & \textbf{0.72} & \textbf{0.82} & \textbf{92.86} \\
 & \textbf{SAP (Ours)} &  & 0.94 & 0.86 & 97.72 & 0.89 & 0.83 & 94.40 & 0.82 & 0.77 & 86.91 \\
\cmidrule{1-12}
\multirow{4}{*}{DocVQA} & EOS-Adaptive & \multirow{4}{*}{0.61} & 0.88 & 0.56 & 90.94 & 0.81 & 0.48 & 78.87 & 0.71 & 0.34 & 56.27 \\
 & Random &  & 0.87 & 0.53 & 86.65 & 0.79 & 0.48 & 77.50 & 0.70 & 0.40 & 65.72 \\
 & Cluster &  & 0.84 & 0.56 & 91.80 & 0.73 & 0.49 & 80.08 & 0.60 & 0.43 & 69.62 \\
 & \textbf{SAP (Ours)} &  & \textbf{0.94} & \textbf{0.60} & \textbf{98.36} & \textbf{0.89} & \textbf{0.59} & \textbf{96.76} & \textbf{0.82} & \textbf{0.54} & \textbf{87.69} \\
\cmidrule{1-12}
\multirow{4}{*}{InfoVQA} & EOS-Adaptive & \multirow{4}{*}{0.92} & 0.84 & 0.86 & 93.08 & 0.75 & 0.77 & 83.86 & 0.64 & 0.60 & 65.54 \\
 & Random &  & 0.87 & 0.88 & 95.70 & 0.80 & 0.85 & 92.27 & 0.72 & 0.78 & 84.40 \\
 & Cluster &  & 0.83 & 0.89 & 96.39 & 0.71 & 0.83 & 90.20 & 0.59 & 0.74 & 80.42 \\
 & \textbf{SAP (Ours)} &  & \textbf{0.92} & \textbf{0.90} & \textbf{97.94} & \textbf{0.86} & \textbf{0.86} & \textbf{93.93} & \textbf{0.78} & \textbf{0.79} & \textbf{86.20} \\
\cmidrule{1-12}
\multirow{4}{*}{Shift Project} & EOS-Adaptive & \multirow{4}{*}{0.89} & 0.85 & 0.87 & 97.81 & 0.79 & 0.81 & 90.16 & 0.72 & 0.70 & 77.83 \\
 & Random &  & 0.88 & 0.84 & 94.09 & 0.81 & 0.78 & 87.01 & 0.73 & 0.65 & 72.93 \\
 & Cluster &  & 0.81 & 0.82 & 91.32 & 0.70 & 0.72 & 80.96 & 0.59 & 0.58 & 65.31 \\
 & \textbf{SAP (Ours)} &  & \textbf{0.90} & \textbf{0.90} & \textbf{100.93} & \textbf{0.84} & \textbf{0.86} & \textbf{96.05} & \textbf{0.78} & \textbf{0.83} & \textbf{92.44} \\
\cmidrule{1-12}
\multirow{4}{*}{SynDocQA-AI} & EOS-Adaptive & \multirow{4}{*}{0.99} & 0.84 & 0.98 & 98.39 & 0.77 & 0.90 & 90.67 & 0.69 & 0.78 & 78.97 \\
 & Random &  & 0.89 & 0.98 & 98.77 & 0.82 & 0.95 & 96.24 & 0.74 & 0.89 & 89.56 \\
 & Cluster &  & 0.81 & 0.97 & 97.90 & 0.71 & 0.92 & 93.11 & 0.60 & 0.90 & 90.43 \\
 & \textbf{SAP (Ours)} &  & \textbf{0.87} & \textbf{0.98} & \textbf{99.00} & \textbf{0.81} & \textbf{0.97} & \textbf{98.00} & \textbf{0.75} & \textbf{0.94} & \textbf{94.63} \\
\cmidrule{1-12}
\multirow{4}{*}{SynDocQA-Energy} & EOS-Adaptive & \multirow{4}{*}{0.96} & 0.85 & 0.91 & 94.74 & 0.78 & 0.86 & 89.37 & 0.70 & 0.72 & 74.87 \\
 & Random &  & 0.89 & 0.94 & 98.18 & 0.82 & 0.90 & 93.52 & 0.73 & 0.82 & 85.64 \\
 & Cluster &  & \textbf{0.81} & \textbf{0.95} & \textbf{98.59} & \textbf{0.71} & \textbf{0.91} & \textbf{94.28} & 0.60 & 0.84 & 87.15 \\
 & \textbf{SAP (Ours)} &  & 0.89 & 0.94 & 97.76 & 0.83 & 0.90 & 93.56 & \textbf{0.76} & \textbf{0.87} & \textbf{91.00} \\
\cmidrule{1-12}
\multirow{4}{*}{SynDocQA-Gov} & EOS-Adaptive & \multirow{4}{*}{0.97} & 0.85 & 0.94 & 97.18 & 0.78 & 0.91 & 93.97 & 0.71 & 0.79 & 81.78 \\
 & Random &  & 0.89 & 0.94 & 96.98 & 0.82 & 0.91 & 93.99 & 0.74 & 0.86 & 88.76 \\
 & Cluster &  & 0.82 & 0.95 & 98.09 & 0.71 & 0.92 & 94.41 & 0.60 & 0.87 & 89.15 \\
 & \textbf{SAP (Ours)} &  & \textbf{0.87} & \textbf{0.96} & \textbf{98.51} & \textbf{0.82} & \textbf{0.94} & \textbf{97.33} & \textbf{0.75} & \textbf{0.89} & \textbf{92.06} \\
\cmidrule{1-12}
\multirow{4}{*}{SynDocQA-Health} & EOS-Adaptive & \multirow{4}{*}{0.98} & 0.85 & 0.97 & 99.12 & 0.78 & 0.93 & 94.61 & 0.71 & 0.82 & 83.81 \\
 & Random &  & 0.89 & 0.97 & 98.68 & 0.82 & 0.95 & 96.89 & 0.74 & 0.90 & 91.56 \\
 & Cluster &  & 0.81 & 0.97 & 98.64 & 0.70 & 0.90 & 92.10 & 0.60 & 0.87 & 88.77 \\
 & \textbf{SAP (Ours)} &  & \textbf{0.88} & \textbf{0.98} & \textbf{99.62} & \textbf{0.82} & \textbf{0.97} & \textbf{99.43} & \textbf{0.75} & \textbf{0.93} & \textbf{94.90} \\
\cmidrule{1-12}
\multirow{4}{*}{TabFQuAD} & EOS-Adaptive & \multirow{4}{*}{0.96} & 0.89 & 0.92 & 96.78 & 0.82 & 0.88 & 92.55 & 0.74 & 0.83 & 86.60 \\
 & Random &  & 0.90 & 0.94 & 98.46 & 0.84 & 0.92 & 96.08 & 0.76 & 0.89 & 92.70 \\
 & Cluster &  & 0.85 & 0.94 & 98.16 & \textbf{0.77} & \textbf{0.93} & \textbf{97.64} & \textbf{0.67} & \textbf{0.90} & \textbf{94.70} \\
 & \textbf{SAP (Ours)} &  & \textbf{0.93} & \textbf{0.94} & \textbf{98.75} & 0.88 & 0.92 & 96.74 & 0.79 & 0.87 & 91.29 \\
\cmidrule{1-12}
\multirow{4}{*}{TAT-DQA} & EOS-Adaptive & \multirow{4}{*}{0.79} & 0.90 & 0.69 & 87.76 & 0.82 & 0.59 & 75.25 & 0.72 & 0.44 & 55.76 \\
 & Random &  & 0.88 & 0.69 & 87.71 & 0.81 & 0.61 & 77.54 & 0.72 & 0.52 & 65.44 \\
 & Cluster &  & 0.81 & 0.71 & 90.05 & 0.71 & 0.67 & 85.09 & 0.61 & 0.58 & 73.87 \\
 & \textbf{SAP (Ours)} &  & \textbf{0.94} & \textbf{0.75} & \textbf{95.57} & \textbf{0.89} & \textbf{0.71} & \textbf{89.51} & \textbf{0.82} & \textbf{0.62} & \textbf{78.33} \\
\bottomrule
\end{tabular}
}
\caption{\textbf{Detailed Performance: Jina Embeddings v4 on ViDoRe v1.} Per-dataset NDCG@5 retention with SAP using the SR-guided window . Best per column in \textbf{bold}.}
\label{tab:v1_jina}
\end{table*}
\clearpage

\begin{table*}[p]
\centering
\scriptsize
\setlength{\tabcolsep}{1.8pt}
\renewcommand{\arraystretch}{1.1}
\resizebox{0.9\textwidth}{!}{
\begin{tabular}{l l c ccc ccc ccc}
\toprule
\multirow{2}{*}{\textbf{Dataset}} & \multirow{2}{*}{\textbf{Method}} & \textbf{Upper Bound} & \multicolumn{3}{c}{$\boldsymbol{\gamma=0.20}$} & \multicolumn{3}{c}{$\boldsymbol{\gamma=0.10}$} & \multicolumn{3}{c}{$\boldsymbol{\gamma=0.05}$} \\
\cmidrule(lr){3-3} \cmidrule(lr){4-6} \cmidrule(lr){7-9} \cmidrule(lr){10-12}
& & \textbf{Full NDCG@5} & \textbf{S.Ret} & \textbf{NDCG@5} & \textbf{\%} & \textbf{S.Ret} & \textbf{NDCG@5} & \textbf{\%} & \textbf{S.Ret} & \textbf{NDCG@5} & \textbf{\%} \\
\midrule
\multirow{4}{*}{Biomed Lec.} & EOS-Adaptive & \multirow{4}{*}{0.57} & 0.88 & 0.50 & 87.16 & 0.82 & 0.44 & 77.73 & 0.76 & 0.39 & 68.66 \\
 & Random &  & 0.93 & 0.54 & 94.77 & 0.89 & 0.52 & 90.71 & \textbf{0.82} & \textbf{0.47} & \textbf{82.12} \\
 & Cluster &  & 0.92 & 0.55 & 95.89 & 0.86 & 0.51 & 88.92 & 0.78 & 0.46 & 80.45 \\
 & \textbf{SAP (Ours)} &  & \textbf{0.94} & \textbf{0.55} & \textbf{96.20} & \textbf{0.89} & \textbf{0.53} & \textbf{93.70} & 0.82 & 0.46 & 80.57 \\
\cmidrule{1-12}
\multirow{4}{*}{Econ. Reports} & EOS-Adaptive & \multirow{4}{*}{0.51} & 0.84 & 0.44 & 86.34 & 0.78 & 0.42 & 83.42 & \textbf{0.71} & \textbf{0.41} & \textbf{80.55} \\
 & Random &  & 0.88 & 0.45 & 88.39 & 0.81 & 0.42 & 82.07 & 0.74 & 0.39 & 76.68 \\
 & Cluster &  & 0.87 & 0.43 & 84.69 & 0.77 & 0.36 & 70.69 & 0.64 & 0.27 & 52.20 \\
 & \textbf{SAP (Ours)} &  & \textbf{0.93} & \textbf{0.47} & \textbf{92.83} & \textbf{0.86} & \textbf{0.46} & \textbf{90.61} & 0.77 & 0.40 & 77.77 \\
\cmidrule{1-12}
\multirow{4}{*}{ESG (Multi)} & EOS-Adaptive & \multirow{4}{*}{0.54} & 0.87 & 0.43 & 78.57 & 0.80 & 0.39 & 71.21 & 0.71 & 0.25 & 46.42 \\
 & Random &  & 0.90 & 0.48 & 87.97 & 0.83 & 0.41 & 75.80 & 0.75 & 0.34 & 62.66 \\
 & Cluster &  & 0.86 & 0.40 & 74.32 & 0.76 & 0.29 & 52.42 & 0.63 & 0.23 & 42.97 \\
 & \textbf{SAP (Ours)} &  & \textbf{0.95} & \textbf{0.50} & \textbf{92.18} & \textbf{0.90} & \textbf{0.48} & \textbf{88.31} & \textbf{0.83} & \textbf{0.42} & \textbf{77.29} \\
\cmidrule{1-12}
\multirow{4}{*}{ESG (Human)} & EOS-Adaptive & \multirow{4}{*}{0.60} & 0.87 & 0.43 & 71.66 & 0.80 & 0.33 & 55.06 & 0.71 & 0.25 & 42.46 \\
 & Random &  & 0.90 & 0.48 & 79.76 & 0.84 & 0.39 & 65.56 & 0.76 & 0.34 & 56.22 \\
 & Cluster &  & 0.86 & 0.47 & 78.70 & 0.76 & 0.38 & 63.95 & 0.64 & 0.25 & 41.85 \\
 & \textbf{SAP (Ours)} &  & \textbf{0.95} & \textbf{0.55} & \textbf{92.53} & \textbf{0.91} & \textbf{0.53} & \textbf{88.88} & \textbf{0.84} & \textbf{0.42} & \textbf{69.93} \\
\bottomrule
\end{tabular}
}
\caption{\textbf{Detailed Performance: ColPali on ViDoRe v2.} Per-dataset NDCG@5 retention with SAP using the SR-guided window . Best per column in \textbf{bold}.}
\label{tab:v2_colpali}
\end{table*}
\clearpage

\begin{table*}[p]
\centering
\scriptsize
\setlength{\tabcolsep}{1.8pt}
\renewcommand{\arraystretch}{1.1}
\resizebox{0.9\textwidth}{!}{
\begin{tabular}{l l c ccc ccc ccc}
\toprule
\multirow{2}{*}{\textbf{Dataset}} & \multirow{2}{*}{\textbf{Method}} & \textbf{Upper Bound} & \multicolumn{3}{c}{$\boldsymbol{\gamma=0.20}$} & \multicolumn{3}{c}{$\boldsymbol{\gamma=0.10}$} & \multicolumn{3}{c}{$\boldsymbol{\gamma=0.05}$} \\
\cmidrule(lr){3-3} \cmidrule(lr){4-6} \cmidrule(lr){7-9} \cmidrule(lr){10-12}
& & \textbf{Full NDCG@5} & \textbf{S.Ret} & \textbf{NDCG@5} & \textbf{\%} & \textbf{S.Ret} & \textbf{NDCG@5} & \textbf{\%} & \textbf{S.Ret} & \textbf{NDCG@5} & \textbf{\%} \\
\midrule
\multirow{4}{*}{Biomed Lec.} & EOS-Adaptive & \multirow{4}{*}{0.54} & 0.84 & 0.46 & 84.58 & 0.77 & 0.43 & 78.49 & 0.67 & 0.38 & 69.43 \\
 & Random &  & 0.90 & 0.52 & 95.09 & 0.83 & 0.49 & 89.89 & 0.76 & 0.45 & 82.05 \\
 & Cluster &  & 0.85 & 0.51 & 93.92 & 0.75 & 0.48 & 87.75 & 0.64 & 0.43 & 79.19 \\
 & \textbf{SAP (Ours)} &  & \textbf{0.93} & \textbf{0.53} & \textbf{96.94} & \textbf{0.87} & \textbf{0.51} & \textbf{93.24} & \textbf{0.81} & \textbf{0.48} & \textbf{88.11} \\
\cmidrule{1-12}
\multirow{4}{*}{Econ. Reports} & EOS-Adaptive & \multirow{4}{*}{0.48} & 0.80 & 0.43 & 89.49 & 0.72 & 0.41 & 85.09 & 0.63 & 0.36 & 75.09 \\
 & Random &  & 0.85 & 0.41 & 85.54 & 0.77 & 0.40 & 82.83 & 0.68 & 0.35 & 73.44 \\
 & Cluster &  & 0.79 & 0.40 & 83.09 & 0.67 & 0.35 & 72.04 & 0.55 & 0.29 & 59.33 \\
 & \textbf{SAP (Ours)} &  & \textbf{0.89} & \textbf{0.45} & \textbf{93.57} & \textbf{0.82} & \textbf{0.42} & \textbf{86.93} & \textbf{0.74} & \textbf{0.38} & \textbf{79.79} \\
\cmidrule{1-12}
\multirow{4}{*}{ESG (Multi)} & EOS-Adaptive & \multirow{4}{*}{0.57} & \textbf{0.89} & \textbf{0.56} & \textbf{98.69} & 0.81 & 0.46 & 81.51 & 0.71 & 0.38 & 67.19 \\
 & Random &  & 0.86 & 0.47 & 82.04 & 0.77 & 0.37 & 65.14 & 0.67 & 0.30 & 52.02 \\
 & Cluster &  & 0.79 & 0.43 & 76.40 & 0.69 & 0.38 & 67.30 & 0.58 & 0.30 & 53.04 \\
 & \textbf{SAP (Ours)} &  & 0.91 & 0.51 & 88.83 & \textbf{0.84} & \textbf{0.49} & \textbf{86.43} & \textbf{0.77} & \textbf{0.41} & \textbf{71.80} \\
\cmidrule{1-12}
\multirow{4}{*}{ESG (Human)} & EOS-Adaptive & \multirow{4}{*}{0.57} & 0.85 & 0.48 & 84.30 & 0.76 & 0.43 & 74.57 & 0.65 & 0.26 & 45.62 \\
 & Random &  & 0.85 & 0.46 & 81.48 & 0.76 & 0.38 & 66.63 & 0.66 & 0.30 & 53.09 \\
 & Cluster &  & 0.79 & 0.46 & 80.81 & 0.69 & 0.44 & 77.91 & 0.58 & 0.31 & 55.18 \\
 & \textbf{SAP (Ours)} &  & \textbf{0.92} & \textbf{0.50} & \textbf{88.29} & \textbf{0.85} & \textbf{0.49} & \textbf{85.68} & \textbf{0.76} & \textbf{0.43} & \textbf{76.13} \\
\bottomrule
\end{tabular}
}
\caption{\textbf{Detailed Performance: ColQwen2 on ViDoRe v2.} Per-dataset NDCG@5 retention with SAP using the SR-guided window . Best per column in \textbf{bold}.}
\label{tab:v2_colqwen}
\end{table*}
\clearpage

\begin{table*}[p]
\centering
\scriptsize
\setlength{\tabcolsep}{1.8pt}
\renewcommand{\arraystretch}{1.1}
\resizebox{0.9\textwidth}{!}{
\begin{tabular}{l l c ccc ccc ccc}
\toprule
\multirow{2}{*}{\textbf{Dataset}} & \multirow{2}{*}{\textbf{Method}} & \textbf{Upper Bound} & \multicolumn{3}{c}{$\boldsymbol{\gamma=0.20}$} & \multicolumn{3}{c}{$\boldsymbol{\gamma=0.10}$} & \multicolumn{3}{c}{$\boldsymbol{\gamma=0.05}$} \\
\cmidrule(lr){3-3} \cmidrule(lr){4-6} \cmidrule(lr){7-9} \cmidrule(lr){10-12}
& & \textbf{Full NDCG@5} & \textbf{S.Ret} & \textbf{NDCG@5} & \textbf{\%} & \textbf{S.Ret} & \textbf{NDCG@5} & \textbf{\%} & \textbf{S.Ret} & \textbf{NDCG@5} & \textbf{\%} \\
\midrule
\multirow{4}{*}{Biomed Lec.} & EOS-Adaptive & \multirow{4}{*}{0.61} & 0.90 & 0.57 & 93.15 & 0.81 & 0.52 & 84.64 & 0.71 & 0.43 & 70.94 \\
 & Random &  & 0.91 & 0.58 & 95.78 & \textbf{0.86} & \textbf{0.56} & \textbf{92.08} & 0.79 & 0.52 & 85.09 \\
 & Cluster &  & 0.86 & 0.57 & 94.07 & 0.79 & 0.55 & 89.88 & 0.70 & 0.48 & 78.95 \\
 & \textbf{SAP (Ours)} &  & \textbf{0.92} & \textbf{0.60} & \textbf{97.94} & 0.87 & 0.55 & 90.96 & \textbf{0.82} & \textbf{0.52} & \textbf{85.72} \\
\cmidrule{1-12}
\multirow{4}{*}{Econ. Reports} & EOS-Adaptive & \multirow{4}{*}{0.55} & 0.87 & 0.52 & 94.91 & 0.79 & 0.48 & 86.98 & \textbf{0.71} & \textbf{0.43} & \textbf{77.51} \\
 & Random &  & 0.88 & 0.49 & 89.38 & 0.81 & 0.44 & 80.53 & 0.74 & 0.40 & 72.68 \\
 & Cluster &  & 0.81 & 0.43 & 77.64 & 0.70 & 0.35 & 62.91 & 0.60 & 0.30 & 54.13 \\
 & \textbf{SAP (Ours)} &  & \textbf{0.91} & \textbf{0.55} & \textbf{99.45} & \textbf{0.84} & \textbf{0.48} & \textbf{88.27} & 0.77 & 0.42 & 76.22 \\
\cmidrule{1-12}
\multirow{4}{*}{ESG (Multi)} & EOS-Adaptive & \multirow{4}{*}{0.53} & 0.91 & 0.45 & 84.56 & 0.84 & 0.43 & 81.14 & 0.74 & 0.34 & 64.37 \\
 & Random &  & 0.89 & 0.45 & 84.20 & 0.82 & 0.40 & 76.15 & 0.73 & 0.30 & 57.03 \\
 & Cluster &  & 0.82 & 0.41 & 78.38 & 0.72 & 0.34 & 63.37 & 0.63 & 0.27 & 51.91 \\
 & \textbf{SAP (Ours)} &  & \textbf{0.93} & \textbf{0.50} & \textbf{94.54} & \textbf{0.88} & \textbf{0.46} & \textbf{87.79} & \textbf{0.82} & \textbf{0.41} & \textbf{77.93} \\
\cmidrule{1-12}
\multirow{4}{*}{ESG (Human)} & EOS-Adaptive & \multirow{4}{*}{0.64} & 0.91 & 0.58 & 90.68 & 0.85 & 0.52 & 82.17 & 0.75 & 0.31 & 48.41 \\
 & Random &  & 0.89 & 0.52 & 81.75 & 0.83 & 0.46 & 73.03 & 0.74 & 0.35 & 55.24 \\
 & Cluster &  & 0.81 & 0.46 & 72.86 & 0.70 & 0.35 & 55.78 & 0.60 & 0.28 & 43.57 \\
 & \textbf{SAP (Ours)} &  & \textbf{0.93} & \textbf{0.58} & \textbf{91.18} & \textbf{0.88} & \textbf{0.55} & \textbf{87.07} & \textbf{0.82} & \textbf{0.42} & \textbf{66.15} \\
\bottomrule
\end{tabular}
}
\caption{\textbf{Detailed Performance: Jina Embeddings v4 on ViDoRe v2.} Per-dataset NDCG@5 retention with SAP using the SR-guided window . Best per column in \textbf{bold}.}
\label{tab:v2_jina}
\end{table*}
\clearpage

\section{Full NDCG Cutoff Results}
\label{app:ndcg_cutoffs}

Tables~\ref{tab:ndcg_cutoffs_v1} and~\ref{tab:ndcg_cutoffs_v2} report NDCG@$k$ retention for $k \in \{1, 5, 10, 50, 100\}$ at the SR-guided window for SAP, on ViDoRe v1 and v2. The spread across cutoffs is below 2.5\% at $\gamma=0.20$ and below 8\% at $\gamma=0.05$, confirming SAP exhibits no cutoff-sensitive degradation.

\begin{table*}[t]
\centering
\small
\setlength{\tabcolsep}{4pt}
\caption{\textbf{Full NDCG cutoffs on ViDoRe v1.} NDCG@$k$ retention (\%) for SAP at the SR-guided window. \textbf{S.Ret} is Score Retention.}
\label{tab:ndcg_cutoffs_v1}
\begin{tabular}{l l c | c c c c c c}
\toprule
\textbf{Model} & \textbf{Method} & $\boldsymbol{\gamma}$ & \textbf{S.Ret} & \textbf{@1} & \textbf{@5} & \textbf{@10} & \textbf{@50} & \textbf{@100} \\
\midrule
\multirow{3}{*}{ColPali} & SAP & 0.2 & 0.94 & 95.5 & 97.5 & 97.5 & 97.8 & 97.8 \\
 & SAP & 0.1 & 0.89 & 91.1 & 93.9 & 94.5 & 95.1 & 95.1 \\
 & SAP & 0.05 & 0.81 & 80.8 & 86.3 & 87.3 & 88.5 & 88.7 \\
\midrule
\multirow{3}{*}{ColQwen2} & SAP & 0.2 & 0.92 & 95.8 & 97.1 & 97.1 & 97.5 & 97.5 \\
 & SAP & 0.1 & 0.86 & 91.4 & 93.2 & 93.7 & 94.2 & 94.3 \\
 & SAP & 0.05 & 0.78 & 81.6 & 85.6 & 86.5 & 87.5 & 87.8 \\
\midrule
\multirow{3}{*}{Jina Embeddings v4} & SAP & 0.2 & 0.91 & 98.2 & 98.4 & 98.6 & 98.7 & 98.7 \\
 & SAP & 0.1 & 0.85 & 93.7 & 95.6 & 95.9 & 96.3 & 96.3 \\
 & SAP & 0.05 & 0.78 & 85.9 & 89.5 & 90.2 & 91.0 & 91.2 \\
\bottomrule
\end{tabular}
\end{table*}

\begin{table*}[t]
\centering
\small
\setlength{\tabcolsep}{4pt}
\caption{\textbf{Full NDCG cutoffs on ViDoRe v2.} NDCG@$k$ retention (\%) for SAP at the SR-guided window. \textbf{S.Ret} is Score Retention.}
\label{tab:ndcg_cutoffs_v2}
\begin{tabular}{l l c | c c c c c c}
\toprule
\textbf{Model} & \textbf{Method} & $\boldsymbol{\gamma}$ & \textbf{S.Ret} & \textbf{@1} & \textbf{@5} & \textbf{@10} & \textbf{@50} & \textbf{@100} \\
\midrule
\multirow{3}{*}{ColPali} & SAP & 0.2 & 0.94 & 93.3 & 93.4 & 94.7 & 95.7 & 96.0 \\
 & SAP & 0.1 & 0.89 & 91.0 & 90.4 & 91.3 & 92.4 & 93.2 \\
 & SAP & 0.05 & 0.81 & 72.2 & 76.4 & 79.2 & 80.9 & 82.3 \\
\midrule
\multirow{3}{*}{ColQwen2} & SAP & 0.2 & 0.91 & 90.1 & 91.9 & 90.8 & 93.2 & 93.6 \\
 & SAP & 0.1 & 0.85 & 85.7 & 88.1 & 86.7 & 88.6 & 89.2 \\
 & SAP & 0.05 & 0.77 & 76.7 & 79.0 & 77.8 & 81.0 & 81.8 \\
\midrule
\multirow{3}{*}{Jina Embeddings v4} & SAP & 0.2 & 0.92 & 94.2 & 95.8 & 95.2 & 96.0 & 96.2 \\
 & SAP & 0.1 & 0.87 & 88.5 & 88.5 & 87.6 & 90.3 & 90.6 \\
 & SAP & 0.05 & 0.81 & 72.9 & 76.5 & 77.8 & 81.7 & 82.8 \\
\bottomrule
\end{tabular}
\end{table*}

\section{Width Robustness of the SR-Guided Window}
\label{app:k_robustness}

\begin{figure*}[h]
    \centering
    \includegraphics[width=\textwidth]{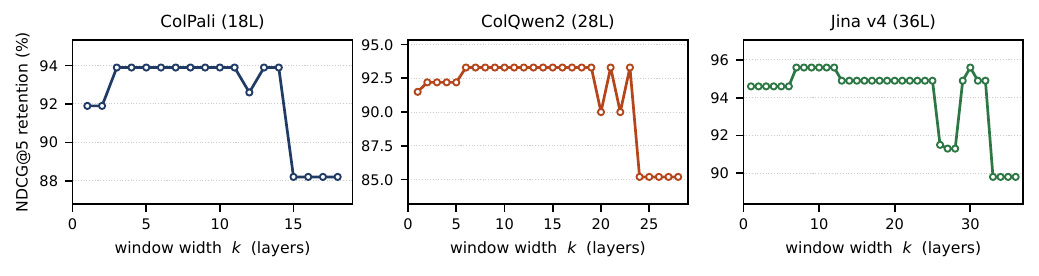}
    \caption{\textbf{Width robustness of the SR-guided window.} NDCG@5 retention at $\gamma=0.10$ (via the nearest 9-window approximation) as the window width $k$ is varied with position fixed by the drop-alignment rule. All three backbones exhibit a broad plateau containing our default $k = \lceil 0.2\,L_{total} \rceil$.}
    \label{fig:k_robustness}
\end{figure*}

Our default fixes the window width at $k = \lceil 0.2\,L_{total} \rceil$ layers. To assess sensitivity to this choice, we hold the window position fixed by the drop-alignment rule  and vary only the width $k$ from $1$ to $L_{total}$, measuring NDCG@5 retention at $\gamma=0.10$ via the nearest 9-window from the brute-force ablation. Figure~\ref{fig:k_robustness} shows the resulting curves for all three backbones. Each backbone exhibits a broad plateau spanning the majority of valid $k$ values, and our default falls comfortably within it (ColPali: $k=4$, ColQwen2: $k=6$, Jina v4: $k=8$). A single-layer window ($k=1$) shows variance due to per-layer centrality noise; overly wide windows dilute the structural signal by mixing aggregation and alignment layers. The fixed default therefore balances both extremes and is not a sensitive hyperparameter.

\section{Calibration Size Robustness}
\label{app:calib_robustness}

The SR-guided procedure has two hyperparameters: the relative window width $\rho$ (analyzed in Appendix~\ref{app:k_robustness}) and the calibration set size $N$. To assess sensitivity to $N$ and to the choice of random sample, we sweep $N \in \{50, 100, 200, 500, 1000\}$ and run the procedure with $5$ different random seeds for each $N$ (samples drawn uniformly from the ColPali training corpus). For each $(N, \mathrm{seed})$ we record the snap window assigned by the drop-alignment rule.

Table~\ref{tab:calib_robustness} summarizes the outcome: ColPali and Jina~v4 converge to a unique window at $N \ge 50$ and $N \ge 100$ respectively; ColQwen2 — the most variance-sensitive backbone — converges to the canonical $60$--$80\%$ window at $N \ge 500$ (one seed drifts to the adjacent $70$--$90\%$ window, which differs by only $1.1\%$ NDCG@5; see Table~\ref{tab:window_ablation}). Our default $N=500$ is therefore sufficient for all three backbones, and doubling to $N=1000$ yields unanimous seed agreement.

\begin{table}[h]
\centering
\small
\setlength{\tabcolsep}{5pt}
\caption{\textbf{Calibration robustness.} SR-guided window selected
by the procedure across calibration sizes $N \in \{50, 100, 200, 500, 1000\}$
and 5 random seeds (samples drawn from the ColPali training corpus).
Each cell reports the most frequent snap window across 5 seeds, with
agreement count in parentheses.}
\label{tab:calib_robustness}
\begin{tabular}{l c c c c c}
\toprule
Backbone & $N{=}50$ & $N{=}100$ & $N{=}200$ & $N{=}500$ & $N{=}1000$ \\
\midrule
ColPali (18L) & 60--80\% (5/5) & 60--80\% (5/5) & 60--80\% (5/5) & 60--80\% (5/5) & 60--80\% (5/5) \\
ColQwen2 (28L) & 60--80\% (3/5) & 80--100\% (2/5) & 70--90\% (3/5) & 60--80\% (4/5) & 60--80\% (5/5) \\
Jina v4 (36L) & 70--90\% (3/5) & 70--90\% (5/5) & 70--90\% (5/5) & 70--90\% (5/5) & 70--90\% (5/5) \\
\bottomrule
\end{tabular}
\end{table}

\section{Full Window Ablation (additional retention ratios)}
\label{app:full_window_ablation}

Tables~\ref{tab:window_ablation_g020} and~\ref{tab:window_ablation_g005} report the full 9-window brute-force ablation at $\gamma=0.20$ and $\gamma=0.05$. The plateau shift with backbone depth (ColPali peaking around 50--80\%, ColQwen2 at 60--80\%, Jina v4 at 70--90\%) is consistent across all three retention ratios.

\begin{table*}[t]
\centering
\small
\setlength{\tabcolsep}{4pt}
\caption{\textbf{Full window ablation at $\gamma=0.20$ (NDCG@5 retention \%) for SAP.} Brute-force best window per row in \textbf{bold}.}
\label{tab:window_ablation_g020}
\begin{tabular}{l l c c c c c c c c c}
\toprule
\textbf{Model} & \textbf{Bench}
& 0--20 & 10--30 & 20--40 & 30--50 & 40--60 & 50--70 & 60--80 & 70--90 & 80--100 \\
\midrule
\multirow{2}{*}{ColPali}
& v1 & 94.1 & 96.3 & 97.2 & 97.6 & \textbf{97.6} & 97.5 & 97.6 & 97.1 & 95.4 \\
& v2 & 89.9 & 93.4 & 94.5 & 93.2 & \textbf{94.7} & 92.9 & 93.9 & 93.4 & 89.6 \\
\cmidrule(lr){1-11}
\multirow{2}{*}{ColQwen2}
& v1 & 93.6 & 95.0 & 95.8 & 95.9 & 96.4 & 96.8 & 97.1 & \textbf{97.2} & 96.8 \\
& v2 & 89.7 & 90.1 & 91.1 & 91.4 & 90.6 & 91.7 & 92.6 & \textbf{92.8} & 88.9 \\
\cmidrule(lr){1-11}
\multirow{2}{*}{Jina v4}
& v1 & 95.0 & 96.0 & 96.5 & 96.4 & 96.9 & 97.6 & 98.2 & \textbf{98.4} & 98.2 \\
& v2 & 88.7 & 89.5 & 93.0 & 93.5 & 95.6 & 94.8 & 93.8 & \textbf{95.8} & 94.3 \\
\bottomrule
\end{tabular}
\end{table*}

\begin{table*}[t]
\centering
\small
\setlength{\tabcolsep}{4pt}
\caption{\textbf{Full window ablation at $\gamma=0.05$ (NDCG@5 retention \%) for SAP.} Brute-force best window per row in \textbf{bold}.}
\label{tab:window_ablation_g005}
\begin{tabular}{l l c c c c c c c c c}
\toprule
\textbf{Model} & \textbf{Bench}
& 0--20 & 10--30 & 20--40 & 30--50 & 40--60 & 50--70 & 60--80 & 70--90 & 80--100 \\
\midrule
\multirow{2}{*}{ColPali}
& v1 & 81.5 & 83.5 & 85.4 & 86.0 & 87.3 & \textbf{88.1} & 86.4 & 85.6 & 81.9 \\
& v2 & 74.3 & 72.2 & 75.0 & 75.1 & 75.5 & \textbf{80.4} & 76.8 & 73.0 & 72.5 \\
\cmidrule(lr){1-11}
\multirow{2}{*}{ColQwen2}
& v1 & 68.6 & 76.4 & 80.4 & 82.0 & 83.2 & 83.6 & 85.5 & \textbf{86.0} & 83.8 \\
& v2 & 68.3 & 71.8 & 76.4 & 76.3 & 76.2 & 77.5 & \textbf{79.6} & 79.3 & 75.2 \\
\cmidrule(lr){1-11}
\multirow{2}{*}{Jina v4}
& v1 & 76.2 & 80.8 & 81.9 & 82.4 & 84.3 & 85.4 & 88.2 & \textbf{89.5} & 87.1 \\
& v2 & 66.9 & 73.4 & 74.8 & 76.3 & 77.1 & 77.3 & \textbf{79.4} & 76.5 & 78.0 \\
\bottomrule
\end{tabular}
\end{table*}

\paragraph{Across two calibration distributions.} Sample size is one axis of
robustness; the calibration \emph{distribution} is another. During development
we ran the SR-guided procedure with two disjoint sources. The first drew 500
pairs as 100 each from five ViDoRe~v1 test subsets (ArxivQA, DocVQA, InfoVQA,
TabFQuAD, TAT-DQA), spanning academic figures, scanned documents, infographics,
tables and financial reports. The second drew 500 pairs from the ColPali
training corpus, which is disjoint from every evaluation split and is the source
used throughout the paper. The two sources select the \emph{identical} window on
all three backbones: the alignment boundary is $l^*=15$, $24$ and $34$, and the
layer sets are $11$--$14$, $18$--$23$ and $26$--$33$, exactly as in
Table~\ref{tab:layer_instantiation}. Together with the size sweep above, this
suggests the window location is a property of the backbone rather than of the
particular document mixture, within the document-image domain. We do not claim
this extends to markedly different visual domains; see Limitations.

\section{Base-Backbone Probe: Per-Region Score Retention}
\label{app:base_probe_values}

Table~\ref{tab:base_probe} gives the values underlying
Figure~\ref{fig:base_probe}. For each backbone and retention ratio we report the
mean Score Retention over the layers inside the SR-guided window
(ColPali $11$--$14$, ColQwen2 $18$--$23$) and over the alignment region that
follows it (ColPali $15$--$17$, ColQwen2 $24$--$27$), with attention read either
from the retrieval model or from its pre-retrieval base backbone. Embeddings,
calibration pairs ($N=500$, seed 42), pruning budget and token positions are
identical within each pair of rows; the attention source is the only variable.

\begin{table}[h]
\centering
\small
\setlength{\tabcolsep}{4pt}
\caption{\textbf{Base-backbone probe.} Mean Score Retention inside the
SR-guided window versus over the alignment region that follows it, with
attention read either from the retrieval-fine-tuned model or from its
pre-retrieval base backbone. Embeddings, calibration pairs ($N=500$, seed 42)
and budget are identical across each pair of rows; only the attention source
changes. \textbf{Drop} is the difference the drop-alignment rule detects.}
\label{tab:base_probe}
\begin{tabular}{l l c c c}
\toprule
\textbf{Backbone, $\gamma$} & \textbf{Attn.\ source} & \textbf{In-win.} & \textbf{After} & \textbf{Drop} \\
\midrule
\multirow{2}{*}{ColPali, $0.10$}
 & fine-tuned & 0.869 & 0.813 & $0.056$ \\
 & base       & 0.874 & 0.855 & $0.019$ \\
\cmidrule(lr){1-5}
\multirow{2}{*}{ColPali, $0.05$}
 & fine-tuned & 0.787 & 0.731 & $0.056$ \\
 & base       & 0.802 & 0.796 & $0.007$ \\
\cmidrule(lr){1-5}
\multirow{2}{*}{ColQwen2, $0.10$}
 & fine-tuned & 0.827 & 0.804 & $0.023$ \\
 & base       & 0.783 & 0.775 & $0.008$ \\
\cmidrule(lr){1-5}
\multirow{2}{*}{ColQwen2, $0.05$}
 & fine-tuned & 0.739 & 0.709 & $0.031$ \\
 & base       & 0.679 & 0.674 & $0.005$ \\
\bottomrule
\end{tabular}
\end{table}

\section{Adaptive Budget Allocation (SAP-Adaptive)}
\label{app:sap_adaptive}

The two dynamic-resolution backbones (ColQwen2, Jina~v4) emit a variable number
of visual tokens per page, so a fixed ratio $\gamma$ already allocates more
patches to denser pages in absolute terms. To test whether reallocating budget
\emph{across} pages helps further, we evaluate SAP-Adaptive: per document we
retain patches whose per-document $z$-score of the SAP centrality exceeds a
threshold, and calibrate that threshold globally with the same quantile
procedure as Appendix~\ref{app:baseline_details} so that corpus-level average
retention still matches the target $\gamma$. Index size is therefore held fixed
and only its distribution over pages changes.

Table~\ref{tab:sap_adaptive} reports ViDoRe v2. The two variants are within
$1.5$ points of each other in eight of nine settings, indicating that the
proportional budget is already close to optimal. The exception is the one the
setting predicts: on ColQwen2's heterogeneous v2 corpora at $\gamma=0.05$,
adaptive allocation gains $3.1$ points. We therefore report SAP-Adaptive as a
deployment option for dynamic-resolution backbones under aggressive
compression, not as a replacement for the fixed-ratio default.

\begin{table}[h]
\centering
\small
\setlength{\tabcolsep}{4pt}
\caption{\textbf{SAP-Adaptive vs.\ SAP-Fixed} on ViDoRe v2 (NDCG@5 retention,
\%). Both variants index the same total number of vectors at each $\gamma$.}
\label{tab:sap_adaptive}
\begin{tabular}{l c c c}
\toprule
\textbf{Backbone} & $\gamma=0.20$ & $\gamma=0.10$ & $\gamma=0.05$ \\
\midrule
ColPali  & 93.5 / 93.4 & 89.8 / 90.4 & 76.3 / 76.4 \\
ColQwen2 & 91.9 / 91.9 & 87.8 / 88.1 & \textbf{82.0} / 79.0 \\
Jina v4  & 97.3 / 95.8 & 89.0 / 88.5 & 77.2 / 76.5 \\
\bottomrule
\multicolumn{4}{l}{\scriptsize Cells read SAP-Adaptive / SAP-Fixed.}\\
\end{tabular}
\end{table}

\section{Visual-to-Visual Attention Ablation}
\label{app:v2v_ablation}

Equation~2 restricts the in-degree column-sum to visual rows. The motivation is
that the indexing prompt is identical for every document in the corpus, so
attention rows originating from those constant text tokens carry an
instruction-specific, document-extrinsic signal; excluding them keeps the score
document-intrinsic, which is what a query-agnostic index-time scorer needs. We
note that under the causal masks of these backbones the visual tokens precede
the instruction tokens, so ``text-to-visual'' attention consists of instruction
rows attending back to image columns; the reverse direction does not exist.

We also test the choice empirically. From the same forward passes over the
500-pair calibration set we compute two centralities: (a) the V2V column-sum of
Equation~2, and (b) an all-to-visual column-sum that additionally includes the
instruction rows. Table~\ref{tab:v2v_ablation} reports window-integrated Score
Retention at three ratios. The two agree to within $0.004$ in all six settings.
A plausible reason is simply quantitative: the prompt contributes fewer than 30
rows against roughly a thousand visual rows, so its share of the column-sum is
small. The top-down prior does not change the patch ranking at these layers, and
we keep the V2V restriction because it is conceptually cleaner.

\begin{table}[h]
\centering
\small
\setlength{\tabcolsep}{4pt}
\caption{\textbf{V2V vs.\ all-to-visual in-degree.} Window-integrated Score
Retention on the 500-pair calibration set, SR-guided window per backbone.}
\label{tab:v2v_ablation}
\begin{tabular}{l c c c c}
\toprule
& \multicolumn{2}{c}{\textbf{ColPali}} & \multicolumn{2}{c}{\textbf{ColQwen2}} \\
\cmidrule(lr){2-3} \cmidrule(lr){4-5}
$\gamma$ & V2V & All-to-vis & V2V & All-to-vis \\
\midrule
0.20 & 0.9453 & 0.9457 & 0.9195 & 0.9204 \\
0.10 & 0.8924 & 0.8917 & 0.8539 & 0.8542 \\
0.05 & 0.8169 & 0.8209 & 0.7673 & 0.7678 \\
\bottomrule
\end{tabular}
\end{table}

\section{Computational Complexity \& Efficiency Analysis}
\label{app:computational_overhead}

A primary concern for any indexing strategy is the additional latency introduced during the document processing phase. In this section, we formally analyze the computational overhead of Structural Anchor Pruning (SAP) and provide empirical benchmarks on the ViDoRe v2 dataset.

\subsection{Theoretical Complexity}
Let $N$ be the number of visual patches (e.g., $1024$ for standard inputs), $L$ the number of transformer layers, and $d$ the hidden dimension.

\paragraph{Backbone Cost.} The computational cost of the standard forward pass is dominated by the self-attention mechanism and feed-forward networks. The complexity for the attention mechanism alone across all layers is $\mathcal{O}(L \cdot N^2 \cdot d)$.

\paragraph{SAP Overhead.}
SAP operates by extracting attention matrices $A^{(l, h)}$ from a subset of layers $\mathcal{L}^*$. The operations required are:
\begin{enumerate}
    \item \textbf{Extraction:} Accessing attention logits (effectively zero FLOPs, bounded by memory bandwidth).
    \item \textbf{Aggregation (In-Degree):} Summing columns of the attention matrix. For a selected layer set $|\mathcal{L}^*|$ and heads $H$, the complexity is $C_{SAP} = \mathcal{O}(|\mathcal{L}^*| \cdot H \cdot N^2)$.
\end{enumerate}

Comparing the two, the ratio of SAP overhead to the attention computation is approximately:
\begin{equation}
    \frac{C_{SAP}}{C_{Attn}} \approx \frac{|\mathcal{L}^*| \cdot H \cdot N^2}{L \cdot H \cdot N^2 \cdot d} = \frac{|\mathcal{L}^*|}{L \cdot d}
\end{equation}
For the \texttt{jina-embeddings-v4} model used in our experiments, with $d=1280$, this ratio is exceedingly small ($< 10^{-3}$), implying the theoretical cost is negligible.

\subsection{Empirical Benchmarks}
\label{sec:empirical_latency}

To validate our theoretical analysis, we conducted a rigorous latency benchmark using the \textbf{ViDoRe v2} dataset (subsets: \textit{esg\_reports, biomedical\_lectures, economics\_reports}).
Experiments were performed on a single \textbf{NVIDIA H200 (141GB)} GPU. The backbone model is \texttt{jina-embeddings-v4} (hidden size $d=1280$).

We measure the \textbf{Pruning Latency} (time taken to compute masks and select tokens) and compare it against the \textbf{Full Forward Pass} time. The results are summarized in Table~\ref{tab:latency_benchmark}.

\begin{table}[h]
\centering
\small
\begin{tabular}{l c r}
\toprule
\textbf{Method} & \textbf{Avg Time (ms)} & \textbf{Overhead (+ \%)} \\
\midrule
\textit{Full Forward Pass} & 206.05 & \textit{(Baseline)} \\
\midrule
Random & 0.04 & +0.02\% \\
Adaptive-EOS & 0.08 & +0.04\% \\
\textbf{SAP (Ours)} & \textbf{0.06} & \textbf{+0.03\%} \\
Cluster-Merge & 12.08 & +5.86\% \\
\bottomrule
\end{tabular}
\caption{\textbf{Efficiency Benchmark on ViDoRe v2.} Latency is the time per page for the pruning operation (mask generation) only. \textbf{Overhead} is calculated relative to the Full Forward Pass time ($206.05$ ms). SAP introduces negligible overhead ($<0.03\%$), whereas clustering-based methods incur a significant penalty ($\approx 6\%$).}
\label{tab:latency_benchmark}
\end{table}

\paragraph{Results Analysis.}
As shown in Table~\ref{tab:latency_benchmark}, the Full Forward Pass requires approximately $206$ms per page.
\begin{itemize}
    \item \textbf{Negligible Overhead:} SAP requires only $0.06$ms per page, corresponding to $\mathbf{\sim 0.03\%}$ overhead relative to the model inference. In a real-world pipeline, this is imperceptible.
    \item \textbf{Comparison to Clustering:} Iterative methods like \textit{Cluster-Merge} (K-Means) are significantly slower, taking $\approx 12$ms per page. While feasible, this represents a $\sim 200\times$ slowdown compared to SAP and adds nearly $6\%$ to the total indexing time.
    \item \textbf{Comparison to Random:} SAP achieves comparable speed to \textit{Random} selection ($0.04$ms) while providing the semantic benefits detailed in Section~\ref{sec:comprehensive_eval}.
\end{itemize}

These results confirm that SAP is a highly scalable solution suitable for high-throughput Visual RAG systems processing millions of documents.

\subsection{Peak Memory During Indexing}
\label{app:memory_benchmark}

Reading attention from a window of layers raises an obvious concern: does
materializing those maps blow up peak memory relative to final-layer EOS
extraction? Two implementation choices keep it bounded. First, SAP reads
attention only from the LLM decoder, so the vision tower stays on SDPA; its
pre-merge patch sequence is roughly four times longer and eager attention there
would dominate. Second, within the decoder each layer's map is reduced to its
column-sum as soon as it is produced and the full tensor is freed before the
next layer runs, so the transient cost is $\mathcal{O}(N^2)$ rather than
$\mathcal{O}(L \cdot N^2)$.

Table~\ref{tab:memory_benchmark} measures peak allocated GPU memory during
indexing on ColQwen2, the dynamic-resolution backbone with the longest
sequences, sweeping the visual-token count through the processor's pixel limit.
The naive strategy of retaining all 28 layers via \texttt{output\_attentions=True}
crosses 32~GB at roughly 5.5k visual tokens and exhausts a 71~GB device beyond
roughly 12k. The streaming implementation adds $0.6$~GB over the plain forward
pass at 3k visual tokens, already well beyond the page sizes in our experiments
($1{,}024$ for ColPali, under 1k for ColQwen2 at default processor settings),
and remains feasible to the processor's maximum sequence length. Final-layer
EOS extraction is the same code path with a single layer, so window extraction
and EOS extraction cost about the same at indexing time.

\begin{table}[h]
\centering
\small
\setlength{\tabcolsep}{3.5pt}
\caption{\textbf{Peak GPU memory during indexing} (ColQwen2, single
NVIDIA H200 MIG 3g.71gb, 69.8~GB usable). (a) plain SDPA forward, no attention
extraction; (b) all 28 decoder layers retained via
\texttt{output\_attentions=True}; (c) our streaming extraction.}
\label{tab:memory_benchmark}
\begin{tabular}{r c c c}
\toprule
\textbf{Visual tok.} & \textbf{(a) SDPA} & \textbf{(c) Streaming} & \textbf{(b) Naive} \\
\midrule
$3{,}025$  & 4.9 GB & 5.4 GB  & 11.0 GB \\
$6{,}084$  & 5.5 GB & 8.8 GB  & 37.0 GB \\
$12{,}100$ & 6.8 GB & 21.6 GB & OOM \\
$16{,}641$ & 7.8 GB & 36.7 GB & OOM \\
\bottomrule
\end{tabular}
\end{table}

\subsection{End-to-End Storage and Retrieval Benchmark}
\label{app:storage_latency}

To complement the per-page pruning-latency analysis above, we measure the
deployment-time index footprint and retrieval throughput on a real
corpus. Setup: ColPali on the ViDoRe~v2 union ($3{,}006$ documents,
$1{,}152$ queries). For each compression ratio $\gamma$, we (1) apply SAP
top-$\lceil \gamma N \rceil$ pruning per document, (2) stack the pruned
embeddings into a single $[\text{docs}, k, d]$ on-disk index, (3) reload
the index into GPU memory, and (4) time end-to-end brute-force MaxSim
retrieval (similarity + per-doc max aggregation + top-5 ranking) over all
queries on a single V100-32GB GPU. Five warmup queries precede each
measurement run; the table below excludes them.

\begin{table}[h]
\centering
\small
\setlength{\tabcolsep}{4pt}
\caption{\textbf{End-to-end storage and retrieval benchmark.} ColPali on
the ViDoRe~v2 union ($3{,}006$ documents, $1{,}152$ queries). Index size
measured on disk; query latency is the mean over all queries
(brute-force MaxSim, V100-32GB). NDCG@5 retention from
Table~\ref{tab:global_results}.}
\label{tab:storage_latency}
\begin{tabular}{c r r r r r}
\toprule
$\gamma$ & Size (MB) & Load (ms) & Lat (ms) & QPS & NDCG@5 \\
\midrule
1.00 & 751.5 & 537 & 6.12 & 163  & 100\% \\
0.20 & 150.4 &  90 & 1.37 & 731  & 93\% \\
0.10 &  74.9 &  51 & 0.78 & 1290 & 90\% \\
0.05 &  37.4 &  28 & 0.47 & 2114 & 76\% \\
\bottomrule
\end{tabular}
\end{table}

At $\gamma=0.10$, SAP compresses storage by $10\times$ and accelerates
retrieval by $7.9\times$; at $\gamma=0.05$, by $20\times$ and $13\times$
respectively. The per-query latency scales roughly as $O(\gamma \cdot
N_{\text{docs}})$ because the dominant cost is the
$\mathrm{query}\!\times\!\mathrm{index}$ matmul, whose cost is linear in
the total patch count. These measured speedups extrapolate directly to
million-scale Visual RAG deployments under the same hardware.

\section{Comparison with Trained Method (Full Table)}
\label{app:trained_comparison}

We benchmark SAP against \textbf{Light-ColPali}~\cite{ma-etal-2025-towards-storage}, a state-of-the-art method that requires supervised training to merge visual tokens. Table~\ref{tab:trained_comparison} reports NDCG@5 and retention \% relative to the full-model upper bound. SAP uses our SR-guided window; the compression factor maps to our retention ratio as $\gamma \approx 1/\text{factor}$, and we report SAP at the closest available $\gamma$. Note that full-model upper bounds and the set of evaluation datasets differ between this work and Light-ColPali; the comparison is therefore indicative, which is why we report retention relative to each method's own upper bound.

\begin{table}[h]
\centering
\small
\setlength{\tabcolsep}{3.5pt}
\caption{\textbf{Training-Free vs.\ Trained Compression.}
SAP vs.\ Light-ColPali~\cite{ma-etal-2025-towards-storage} (trained token merging) on ColPali and ColQwen2. Subscripts denote \% retention of the corresponding full model. We exclude a Jina v4 + Light-Merging baseline because Light-ColPali releases trained weights only for the ColPali and ColQwen2 backbones.}
\label{tab:trained_comparison}
\resizebox{\columnwidth}{!}{
\begin{tabular}{l l c c c c}
\toprule
\multirow{2}{*}{\textbf{Backbone}} & \multirow{2}{*}{\textbf{Method}} & \textbf{Training} & \multicolumn{3}{c}{\textbf{Compression Factor}} \\
\cmidrule(lr){4-6}
& & \textbf{Req.} & \textbf{$4\times$} & \textbf{$9\times$} & \textbf{$25\times$} \\
\midrule
\multirow{2}{*}{\textbf{ColPali}}
& Light-ColPali$^\dagger$           & Yes         & $0.75_{\scriptscriptstyle 98.4}$           & $\mathbf{0.75}_{\scriptscriptstyle \mathbf{97.8}}$ & $\mathbf{0.73}_{\scriptscriptstyle \mathbf{95.2}}$ \\
& \textbf{SAP (Ours)}               & \textbf{No} & $\mathbf{0.83}_{\scriptscriptstyle \mathbf{98.2}}$ & $0.80_{\scriptscriptstyle 93.9}$                    & $0.74_{\scriptscriptstyle 86.3}$ \\
\midrule
\multirow{2}{*}{\textbf{ColQwen2}}
& Light-ColQwen2$^\dagger$           & Yes         & $\mathbf{0.81}_{\scriptscriptstyle \mathbf{99.0}}$ & $\mathbf{0.80}_{\scriptscriptstyle \mathbf{98.2}}$ & $\mathbf{0.78}_{\scriptscriptstyle \mathbf{96.3}}$ \\
& \textbf{SAP (Ours)}               & \textbf{No} & $0.86_{\scriptscriptstyle 97.7}$                    & $0.82_{\scriptscriptstyle 93.2}$                    & $0.75_{\scriptscriptstyle 85.6}$ \\
\bottomrule
\multicolumn{6}{l}{\scriptsize $^\dagger$ Results cited from Light-ColPali~\cite{ma-etal-2025-towards-storage}.}\\
\end{tabular}
}
\end{table}

\section{SR--NDCG Correlation: Scatter Plot}
\label{app:sr_ndcg_scatter}

\begin{figure}[h]
    \centering
    \includegraphics[width=\linewidth]{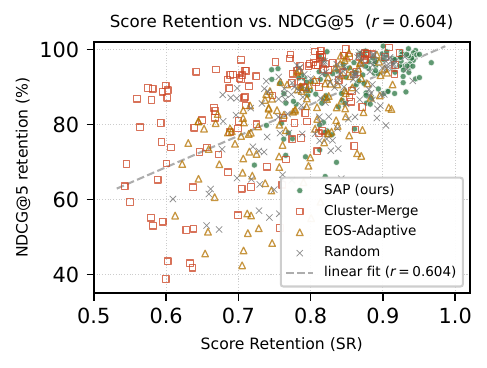}
    \caption{\textbf{Score Retention vs.\ NDCG@5 retention.} Each point is one (model, dataset, method, $\gamma$) configuration. SAP (solid markers) consistently occupies the high-fidelity region; baselines (hollow markers) exhibit substantially more scatter. The moderate correlation ($r \approx 0.60$) reflects that SR measures per-pair fidelity while NDCG measures corpus-level ranking; the two are complementary, not interchangeable.}
    \label{fig:retention_correlation}
\end{figure}

Figure~\ref{fig:retention_correlation} reports the SR--NDCG@5 retention scatter across all (model, dataset, method, $\gamma$) configurations. SAP (solid markers) consistently occupies the high-fidelity, high-NDCG region; baselines (hollow markers) exhibit substantially more scatter. The Pearson correlation of $r \approx 0.60$ confirms a meaningful but imperfect relationship between SR and NDCG, consistent with their distinct definitions (Section~\ref{sec:oracle_protocol}).

\end{document}